\pdfoutput=1

\documentclass[11pt]{article}

\usepackage[]{acl}

\usepackage{mathptmx}
\usepackage{latexsym}
\usepackage{hyperref}
\hypersetup{colorlinks,allcolors=black}

\usepackage[export]{adjustbox}
\usepackage{multirow}
\usepackage[T1]{fontenc}

\usepackage[utf8]{inputenc}

\usepackage{numprint}
\usepackage{enumitem}
\usepackage{graphicx}
\usepackage{amssymb}
\usepackage{pifont}
\usepackage{tabularx}
\usepackage{booktabs}
\usepackage[american]{babel}
\usepackage[autostyle=true]{csquotes}
\usepackage{caption}
\usepackage{subcaption}

\usepackage{microtype}
\usepackage{color}

\title{You Shall Know a Tool by the Traces it Leaves: The Predictability of Sentiment Analysis Tools}

\author{Daniel Baumartz \and Mevlüt Bagci \and Alexander Henlein \and Maxim Konca \\ {\bf \and Andy Lücking \and Alexander Mehler} \\
  Text Technology Lab \\
  Goethe University Frankfurt, Germany \\
  \texttt{\{baumartz,bagci,henlein,konca,luecking,mehler\}@em.uni-frankfurt.de}}

\newcommand{\cmark}{\ding{51}}
\newcommand{\kreis}[1]{{\textcircled{\small{#1}}}}
\newcommand{\quadneu}[1]{{\fbox{\small{#1}}}}
\begin{document}
\maketitle
\begin{abstract}
If sentiment analysis tools were valid classifiers, one would expect them to provide comparable results for sentiment classification on different kinds of corpora and for different 
languages.
In line with results of previous studies we show that sentiment analysis tools disagree on the same dataset.
Going beyond previous studies we show that the sentiment tool used for sentiment annotation can even be predicted from its outcome, revealing an \emph{algorithmic bias} of sentiment analysis.
Based on Twitter, Wikipedia and different news corpora from the English, German and French languages, our classifiers separate sentiment tools with an averaged $F_1$-score of $0.89$ (for the English corpora).
We therefore warn against taking sentiment annotations as face value and argue for the need of more and systematic NLP evaluation studies.
\end{abstract}

\section{Introduction}
\label{sec:introduction}
Mining opinions and valuations -- sentiment -- is an important tool in the repertoire of commercial and non-commercial social media analysts \citep{Taboada:2016}.
Educational data mining, for instance, is employed for evaluating microblogging of educational institutions \citep{Kimmons:Veletsianos:Woodward:2017} and the ranking of universities \citep{Abdelrazeq2016}.
Sentiment classification on tweets is part of predicting stock market events \citep{BOLLEN20111}.
Understanding elections refers, among others, to mood as communicated on social platforms \citep{MOHAMMAD2015480}.
All of these approaches have in common that they use single or multiple NLP tools (e.g., from the field of sentiment analysis) to conduct a study in a research area (e.g., computational sociology, psychology, education) outside NLP.
For this purpose, they use tools whose authors usually report F-scores above the corresponding SOTA at the time of publication.
It is a truism that different tools can perform differently for the same task on the same text corpus (due to different training datasets, training data domains, etc.) \citep{King:1996,ortmann2019evaluating,wiegreffe2021teach}.
However, this becomes problematic if the tools are not systematically compared to examine their impact on reported outcomes.
In this case, there is a risk that, for example, a sociological, psychological or educational statement is based on the results of one or a few tools without knowing whether these are valid and whether they are not rather biased.
This risk becomes apparent when different tools provide very different sentiment results for the same data.
As a result of this diversity, the tools can ultimately become recognizable and thus distinguishable from one another through the sentiment analyses they output.
The sentiment analyses are then \textit{algorithmically biased}, so to speak:
one cannot claim to have determined valid sentiment values with a particular tool, but only those for which the biasing algorithm being applied was decisive.
In short, it is not only since the success of neural networks that many scientific disciplines use NLP tools to conduct text-based studies with the aim of substantiating their research \citep[e.g.][]{Thessen:et:al:2012}. 
Our hypothesis is that these tools, although they may produce comparable F-scores in evaluations, tend to produce different distributions of output values that may ultimately cover the entire range of (e.g., of sentiment) values.
This results in the predictability of the tools based on the output patterns they produce.
And the better this predictability, the higher the algorithmic bias caused by the use of these tools.

In this work, we investigate tool predictability in the area of sentiment analysis using corpora from three languages: English, French and German.
We test 9 tools, which, depending on methodology and language results in 9 instances for English and 4 for German and French each.
We consider four areas of test data -- Twitter, Wikipedia, newspaper and Europarl corpus data -- to make our analysis less genre-dependent.
Our basic finding is that tool predictability does indeed exist to a relevant extent: 
In most cases, it is sufficient to examine some statistical moments (e.g., mean and standard deviation) of the sentiment value distribution generated by a tool for the analyzed texts to know which tool it is.
To show this, the paper is organized as follows:
In \autoref{sec:related_work} we present related work on sentiment analysis. 
In \autoref{sec:data_tools} we introduce the data and tools we cover in our experiments.
Section~\ref{sec:experiments} presents our experiments and results, which we discuss in \autoref{sec:discussion}.
We conclude in \autoref{sec:conclusion}.

\section{Background and related work}
\label{sec:related_work}

Twitter is a particularly useful corpus for sentiment studies \citep{feng-etal-2013-twitter}.
However, there is a lot of evidence that sentiment classification is influenced by linguistic and extra-linguistic factors such as the information within an explicit sentiment lexicon \citep{agarwal-etal-2011-sentiment}, (non-)recognition of implicit sentiment \citep{van-hee-etal-2021-exploring}, and specific syntactic constructions \citep{verma-etal-2018-syntactical}, but not by translated tweets \citep{salameh-etal-2015-sentiment,lu-etal-2011-joint}.
The choice of the tool used for assessing sentiment has a major effect on the outcome of any classification, too.
A systematic comparison of 20 sentiment recognition tools (15 stand-alone, 5 workbenches) on texts from five tweet genres -- Pharma, Retail, Security, Tech, Telco -- has been carried out by \citet{abbasi-etal-2014-benchmarking}. 
Crucially, the authors found out that the tools not only performed differently on average, but also varied for each text genre.
The main error the tools are prone to is ascribed to a lack of recognizing user intentions. 
Little agreement of tools to manually labeled datasets, and even less agreement among the tools themselves, has also been observed with software engineering data \citep{Jongeling:et:al:2015}; 
the best (that is, the least worst) performing tools (NLTK and SentiStrength) even come to disagreeing classifications for different kinds of datasets.
Although combining various sentiment classifiers leads to a wider coverage, it does not necessarily lead to a better F-score \citep{Goncalves:et:al:2014}. 
It has to be kept in mind, however, that any comparisons can be influenced by differences in annotation guidelines and annotators \citep{Mozetic:et:al:2016} or varying decisions on stop word lists \citep{MAYNARD16.188}.
The biases bound up with tool choice are amplified by biases of the language models used by the tools \citep[e.g.][]{Fokkens:et:al:2013} and their size \citep{Bender:et:al:2021}. 
Experimenting with random seeds and stochastic weight averaging (SWA) from the BERT-based ALBERTA model on sentiment data (compared with a check list evaluation), \citet{khurana-etal-2021-emotionally}, e.g., found that the error rate of random seeds is reduced by SWA but the still leaves a margin of instability.
Thus, though important for commercial and non-commercial applications, from recent related work it is known that sentiment classification of tweets and on other kinds of corpora appears to be a fragile affair.
In what follows, we show that we are able to identify sentiment analysis tools based solely on their produced output values -- shedding light on another aspect of the susceptibility of sentiment analysis.
\section{Data and tools}
\label{sec:data_tools}
The following corpora and sentiment tools have been selected for our experiments.

\subsection{Data}
\label{sec:data}
We consider three languages: English (EN), French (FR) and German (DE).
For each, we built three corpora based on tweets taken from Twitter and a selection of Wikipedia and newspaper articles.
In addition, we used the parallel Europarl corpus to compare sentences in these three languages with the same underlying meaning.
\autoref{tab:corpora_stats} in the appendix provides an overview of the corpora.
\begin{description}[style=unboxed,leftmargin=0cm]
    \item[Twitter]
    Using the Full-Archive-Search of their API, we downloaded tweets from Twitter 
    querying for the following hashtags related to the COVID-19 pandemic, that is discussions in the German speaking community: \#Aufschrei, \#allemalneschichtmachen, \#allenichtganzdicht, \#allesdichtmachen, \#lockdownfuerimmer, \#niewiederaufmachen and \#TatortBoykott, as well as the language independent topic about the Sputnik V vaccine: \#SputnikV~\cite{Abrami:Mehler:2021}.
    The tweets were created between 2013-06-03 and 2021-04-24 (see \autoref{tab:corpora_stats} for more details).
    \item[Wikipedia] We randomly selected a sample of about \numprint{20000} Wikipedia 
    articles independently from the English, French and German Wikipedia, based on the 2021-06-20 dump
    .
    \item[News corpora] We analyzed around \numprint{20000} newspaper articles from the New York Times (NYT)~\citep{NYT:2019}
    , the Süddeutsche Zeitung (SDZ)~\citep{SZ:2014} 
    and \numprint{10000} articles from the French 2020 10k Newscrawl by the Leipzip Corpora Collection~\citep{DBLP:conf/lrec/GoldhahnEQ12}. 
    The articles from NYT and SDZ were randomly selected to be evenly distributed by percentage over the years of publication, ranging from 1987 to 2019 in case of the NYT and 1992 to 2014 in case of SDZ.
    \item[Europarl] Using the Europarl~\citep{DBLP:conf/mtsummit/Koehn05} 
    corpus (released 2012 in version 7), we created a parallel mapping of sentences of all three languages.
\end{description}
The corpora have all been preprocessed by spaCy\footnote{\url{https://spacy.io}} (using the small/efficient models) running inside TextImager\footnote{The TextImager platform has since been further developed into the \textit{Docker Unified UIMA Interface} (DUUI)~\cite{Leonhardt:et:al:2023}.}~\cite{Hemati:Uslu:Mehler:2016}, an NLP platform based on Apache UIMA.
We integrated the sentiment tools described in the next section into our platform by mapping their inputs and sentiment output to UIMA.
\subsection{Sentiment Analysis Tools}
\label{sec:tools}
We computed coarse-grained polarity sentiments for all corpora of the three languages using tools in two ways:
Firstly, the entire text (i.e., the full tweet or Wikipedia article) is used as input, and secondly, each of its sentences is processed individually.
To always generate a single sentiment value per text, we followed the recommendation of the tools and average over the sentiments of all sentences.
In this way, we generate two sentiment scores for each language per corpus, per tool and text.
We chose tools that operate differently to capture and compare a broader range.
This includes tools that utilize heuristics build upon lexica as well as model-based algorithms.
To unify the usage of different tools, we map their respective outputs to a sentiment range of $-1$ (negative) to $+1$ (positive).
The following tools are used (for an overview of them, the data on which their sentiment detection is based, and links to the implementations we used in our experiments, see \autoref{tab:tools_used_datasets} in the appendix): 
\begin{description}[style=unboxed,leftmargin=0cm]
    \item[TextBlob] offers two methods for sentiment detection that are lexicon- and model-based.
    The lexicon is being used by a pattern analyzer based on the pattern library~\citep{DBLP:journals/jmlr/SmedtD12}.
    The analyzer produces sentiment polarity values in the range of $-1$ (negative) to $+1$ (positive).
    The model-based method uses a naive Bayes classifier.
    Sentiments are output by classification (\textit{pos} or \textit{neg}) together with two values denoting positive and negative scores in the interval $[0,1]$.
    We use the positive score if the classification yields \textit{pos}, otherwise the negative score multiplied by $-1$ to comply with our sentiment range.
    We use textblob-de, which provides a German sentiment lexicon mainly based on the German Polarity Lexicon~\citep{zora45506}, and textblob-fr for French data.
    \item[Stanza] \citet{qi2020stanza} provide sentiment detection for Chinese, English and German by means of a CNN-based classifier based on \cite{kim-2014-convolutional}.
    It produces a discrete value of $0$, $1$ or $2$ denoting negative, neutral or positive sentiment, allowing for a direct mapping to our schema.
    Multiple datasets are used by the English model, that is the Stanford Sentiment Treebank\footnote{\href{https://github.com/stanfordnlp/sentiment-treebank}{\nolinkurl{https://github.com/stanfordnlp/sentiment-treebank}}}, MELD~\citep{DBLP:conf/acl/PoriaHMNCM19,DBLP:conf/aaai/ZahiriC18}, Sentiment Labelled Sentences Data Set~\citep{DBLP:conf/kdd/KotziasDFS15}, ArguAna TripAdvisor Corpus~\citep{wachsmuth:2014a} and Twitter US Airline Sentiment\footnote{\href{https://www.kaggle.com/crowdflower/twitter-airline-sentiment/data}{\nolinkurl{https://www.kaggle.com/crowdflower/twitter-airline-sentiment/data}}}.
    The German model is trained on the German Tweet corpus SB-10k~\citep{cieliebak-etal-2017-twitter}.
    \item[VADER]
    ~\citep{DBLP:conf/icwsm/HuttoG14} uses a manually validated lexicon based on the sentiment word-banks Linguistic Inquiry Word Count~\citep{DBLP:conf/latech/AndreiDDM14}, General Inquirer~\citep{DBLP:conf/afips/StoneH63} and Affective  Norms  for  English  Words~\citep{bradley1999affective}, enhanced by expressions commonly used in social media posts.
    Its rules are based on five heuristics that incorporate word-order sensitive relationships.
    We use the continuous compound score of VADER as sentiment value, which lies in the interval of $-1$ (negative) to $+1$ (positive).
    To process German text we utilize GerVADER~\citep{DBLP:conf/lwa/TymannLPG19} which provides a new lexicon, mostly based on SentiWS~\citep{DBLP:conf/lrec/RemusQH10} with additions by Langenscheidt\footnote{\url{https://www.langenscheidt.com/jugendwort-des-jahres}} and CoolSlang\footnote{\url{https://www.coolslang.com}}, and updated heuristics.
    VADER-FR is used for French text; it contains a manually translated lexicon.
\end{description}
We use six different single- and multi-language models based on language models like BERT~\citep{DBLP:conf/naacl/DevlinCLT19} and derivates.
All these models produce discrete sentiment values that we mapped to $-1$, $0$, $+1$ and interim values, if needed.
\begin{description}[style=unboxed,leftmargin=0cm,noitemsep,topsep=0cm]
    \item[cardiffnlp/twitter-roberta-base-sentiment]~\citep{barbieri-etal-2020-tweeteval} is trained on English tweets and finetuned on the evaluation framework TweetEval, with sentiment data from SemEval 2017 Task 4 Subtask A~\citep{DBLP:journals/corr/abs-1912-00741}.
    \item[cardiffnlp/twitter-xlm-roberta-base-sentiment]~\citep{barbieri2021xlmt} supports multiple languages. We used it to process English and French texts.
    It is trained on Twitter and finetuned on the UMSAB dataset, which is based on SemEval 2017 Task 4 Subtask A~\citep{DBLP:journals/corr/abs-1912-00741} (EN), SB-10k~\citep{cieliebak-etal-2017-twitter} (DE) and Deft 2017~\citep{oatao19108} (FR).
    \item[finiteautomata/bertweet-base-sentiment\-analysis]~\citep{perez2021pysentimiento} is based on BERTweet~\citep{bertweet}, a language model of English tweets, and trained with TASS 2020 Task 1~\citep{DBLP:conf/sepln/VegaDCAMZCACCM20} and SemEval 2017 Task 4 Subtask A~\citep{DBLP:journals/corr/abs-1912-00741}.
    \item[nlptown/bert-base-multilingual-uncased\-sentiment] was used to detect sentiments in our English and French corpora.
    It is trained on product reviews.
    \item[oliverguhr/german-sentiment-bert] \citet{guhr-EtAl:2020:LREC} trained a model with German texts from various domains including Twitter: PotTS~\citep{DBLP:conf/lrec/Sidarenka16}, SB-10k~\citep{cieliebak-etal-2017-twitter}, GermEval-2017~\citep{wojatzki2017germeval}, Scare~\citep{DBLP:conf/lrec/SangerLKAK16}, leipzig-wikipedia~\citep{DBLP:conf/lrec/GoldhahnEQ12}, crawled data from Filmstarts\footnote{\url{https://www.filmstarts.de}} and HolidayCheck\footnote{\url{https://www.holidaycheck.de}} and Emotions based on experiments with service robots~\citep{guhr-EtAl:2020:LREC}.
    \item[siebert/sentiment-roberta-large-english] \citet{heitmann2020more} provide a binary sentiment detection model for English trained on a total of 15 datasets by
    \citet{blitzer-etal-2007-biographies, pang-lee-2005-seeing, 10.1145/2507157.2507163, speriosu-etal-2011-twitter, HARTMANN201920, maas-etal-2011-learning, nakov-etal-2013-semeval, shamma2009, pangetal2002thumbs}, as well as the Yelp Academic\footnote{\url{https://www.yelp.com/dataset}} and Kaggle sentiment140\footnote{\url{https://www.kaggle.com/kazanova/sentiment140}} datasets.
    Note that while this model cannot produce neutral sentiments, such values can still appear in case of the variant where we consider sentences as inputs and average over them.
\end{description}
\section{Experiments}
\label{sec:experiments}
We evaluate the sentiment values computed for the corpora of \autoref{sec:data} and find a large variance between the different tools.
\begin{figure}
    \centering
    \includegraphics[width=\linewidth]{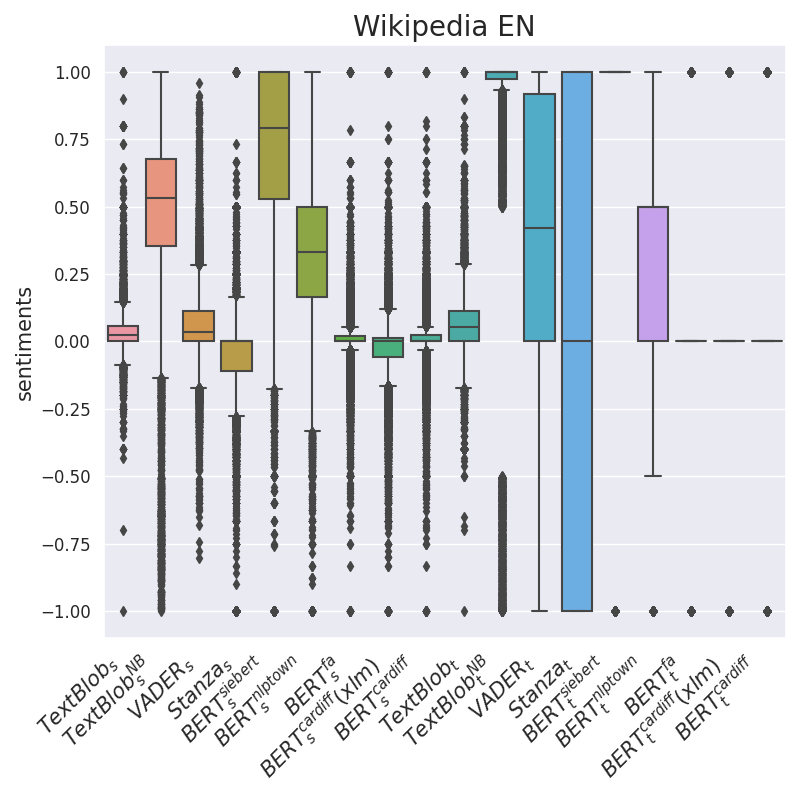}
    \includegraphics[width=\linewidth]{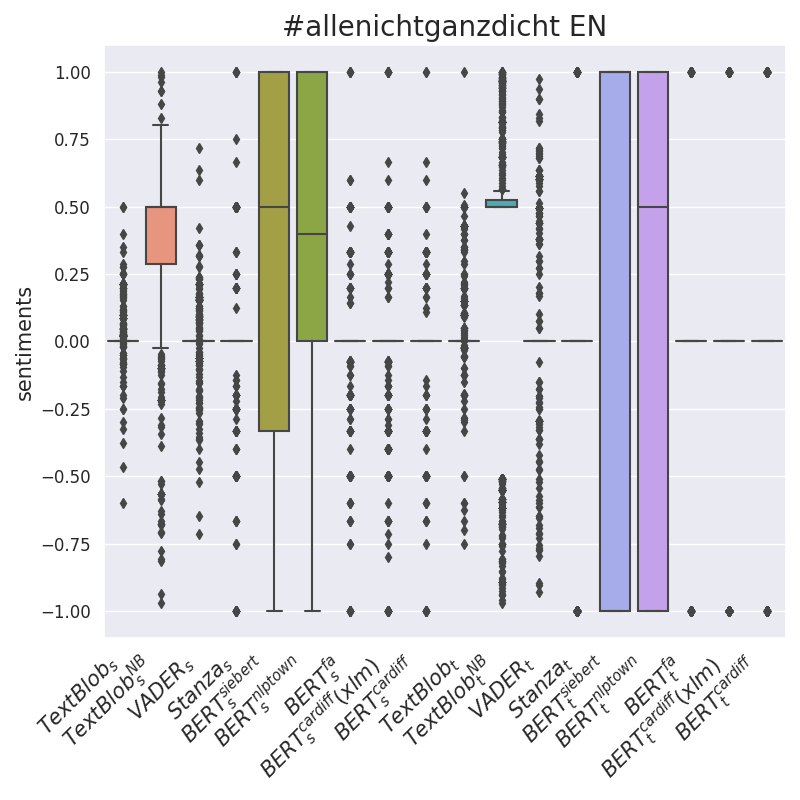}
    \includegraphics[width=\linewidth]{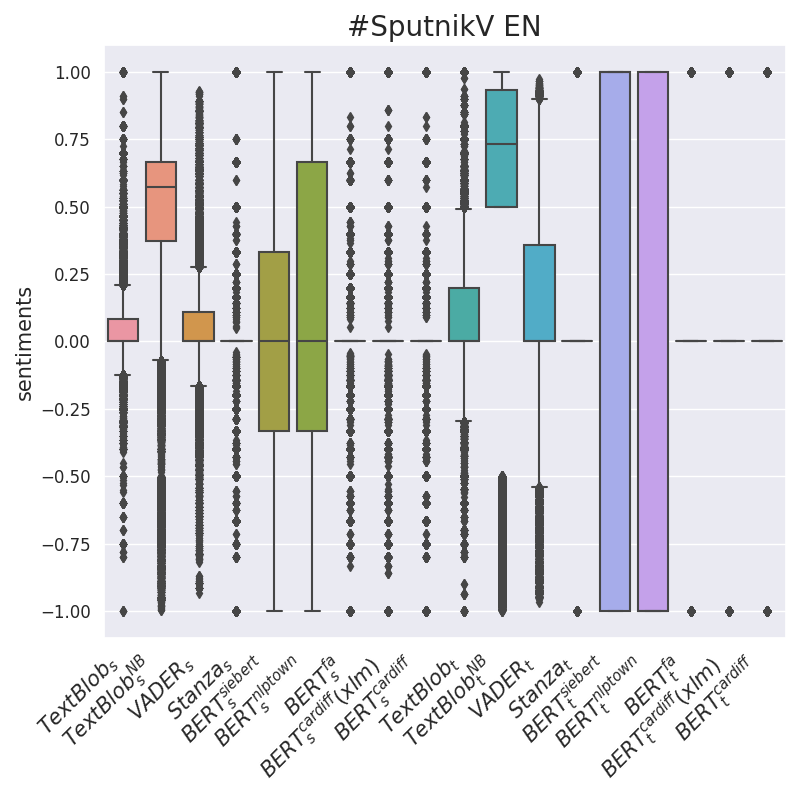}
    \caption{Sentiments of English Wikipedia and Twitter hashtags \#allenichtganzdicht and \#SputnikV.}
    \label{fig:boxplot_enwiki_sputnikv}
\end{figure}
This becomes apparent when the mean and standard deviation are calculated across all sentiments -- see \autoref{fig:boxplot_enwiki_sputnikv} for an overview of the values on the EN Twitter \#SputnikV corpus and the EN Wikipedia.
\autoref{fig:boxplot_defrwiki_sputnikv} in the appendix exemplifies these analyses for DE and FR.
\autoref{fig:dcor_all} shows the distance correlation of the sentiment predictions for all EN corpora except Europarl (see Figure \ref{fig:dcor_all_de_fr_en} in the appendix for DE and FR).
The correlations are low (indicated by light green and white, respectively), suggesting that the tools make rather heterogeneous predictions.
\textit{But does that finding already make them predictable?}

\begin{figure}
    \centering
    \includegraphics[width=\linewidth]{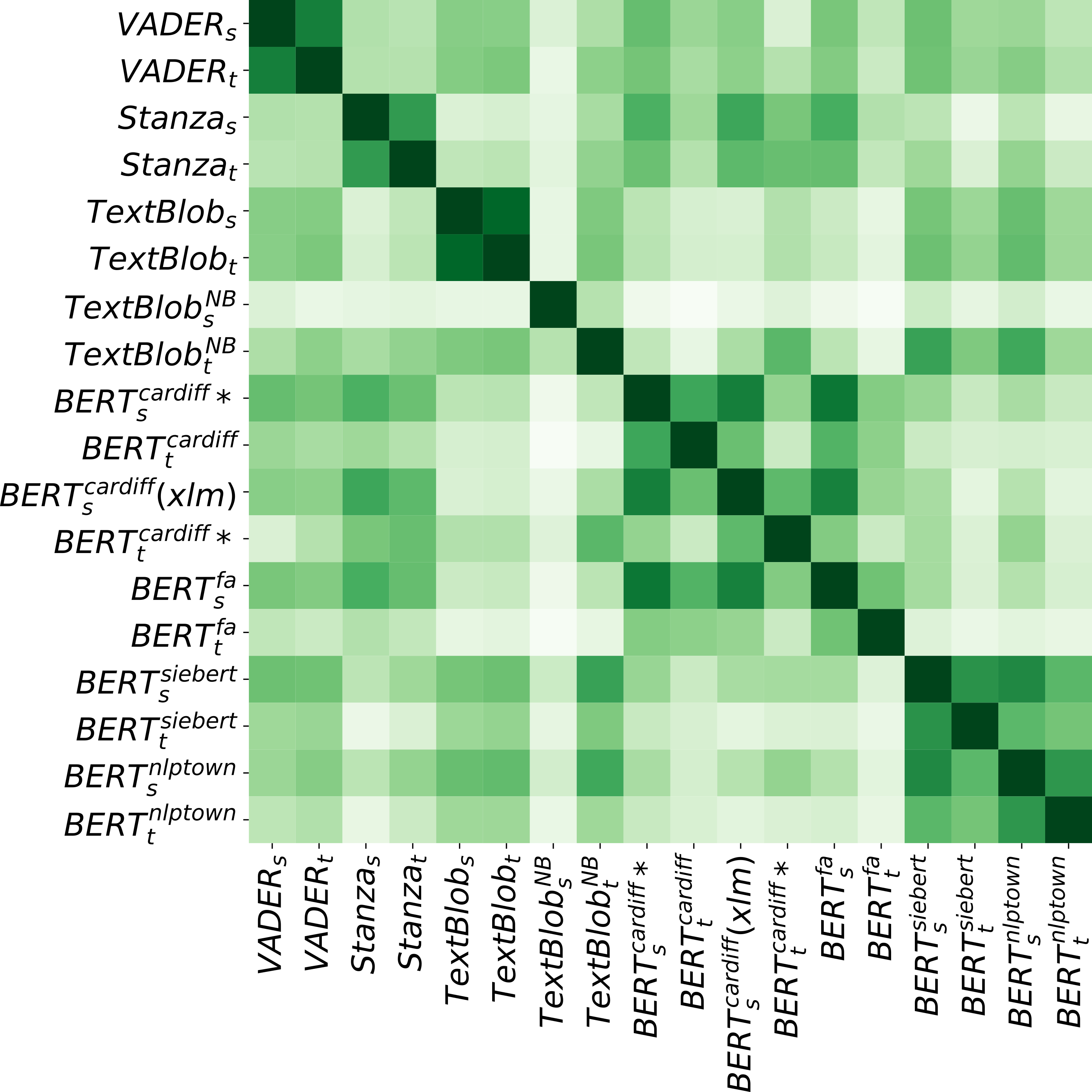}
    \captionsetup{skip=5pt}
    \caption{Distance correlation of sentiment scores for different tools on the EN C3 corpora, with darker color indicating higher correlation.}
    \label{fig:dcor_all}
\end{figure}
\begin{figure}
    \centering
    \includegraphics[width=\linewidth]{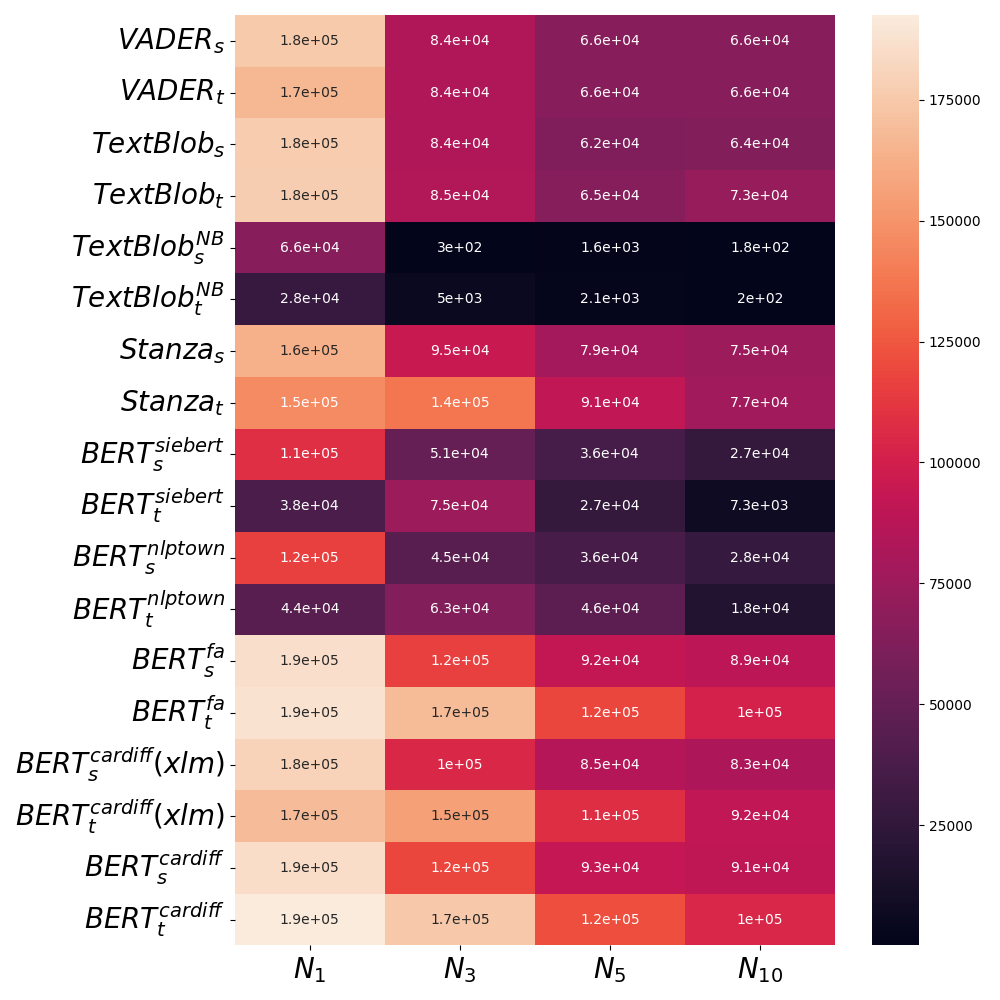}
    \captionsetup{skip=5pt}
    \caption{Per-tool rate of agreement with majority vote for the EN C3 corpus using 4 normalization methods.}
    \label{fig:voting_en}
\end{figure}

\autoref{fig:voting_en} shows how often the tools for EN returned a result equal to a majority vote.
For each document, this vote is calculated using the original sentiment tool results, normalized in four different ways (as described in \autoref{sec:nn}).
As expected, the agreement decreases with more possible output margins due to normalization.
We observe the same output for DE and FR (s.\ \autoref{fig:voting_de_fr} in the appendix).
3 of the 5 transformer models have almost the same agreement rate.
This could indicate that they produce similar analyses. So are they indistinguishable?
We will now show that they are.
Our main hypothesis about tool predictability is that we are able to identify the tool used to perform sentiment analysis by the statistical signature of its output.
To test this hypothesis, we train several classifiers that recognize tools based on this  signature (in terms of \textit{mean}, \textit{std} etc.). 
We use the tools as target labels and perform a binary or multi-class classification, depending on the underlying tool combination.
In addition to experiments that consider all data and tools, we explore several combinations based on the properties of the tools and the domains of our corpora.
\subsection{Neural network classifier}
\label{sec:nn}
Using PyTorch~\citep{NEURIPS2019_9015} 
and scikit-learn~\citep{scikit-learn} 
we train a shallow neural network (NN) with one hidden linear layer, a ReLU activation function and softmax output.
We generate training samples by first randomly collecting subsets of documents. Then we build for each of these subsets and each tool a so called chunk as the multiset of all sentiment values that this tool produces for the documents included in the subset.
Given a tool, we then compute for each chunk 13 different statistical moments using numpy~\citep{2020NumPy-Array} 
(i.e., \texttt{mean}, \texttt{std}, \texttt{var}, \texttt{median}, \texttt{min}, \texttt{max} and the 5-, 10-, 25-, 50-, 75-, 90- and 95-\texttt{percentiles}). 
In this way, for each of the $n$ tools, we obtain a feature vector consisting of $13$ times the number of chunks many elements as input to classification.
We vary chunk size in the range of $50$--$1000$ documents; 
in \autoref{sec:monto_carlo} we additionally show effects of varying the sampling method.
The training dataset is randomly split in a train set (70~\%), a development set and a test set (15~\% each).
Because the tools output different ranges of sentiment, we conducted additional experiments in which we normalized all values and classified the tools into different groups based on their methodology.
We normalize sentiment values in four ways:
firstly, we discretize them by returning \textit{positive} for values $\ge 0.5$, \textit{negative} for $\le -0.5$, and \textit{neutral} otherwise; this results in $N_{1}$.
Secondly, we generate normalization $N_{3}$ according to the membership of sentiment values in one of the three intervals $(-\infty, -0.333]$, $(-0.333, 0.333)$, and $[0.333,\infty)$.
Finally, we divide the ranges of sentiment scores into five ($N_{5}$) or ten ($N_{10}$) equally sized intervals to account for more sentiment gradations.
The tools are independently divided into the following groups:
a group containing all tools that provide discrete sentiment outputs (i.e., the transformer-based models and Stanza), and the group of all remaining tools (i.e., TextBlob and VADER) that output numerical values.
This is to investigate the difficulty of tool prediction when only a few different sentiment values are produced, as is the case with discrete outputs.
To prevent that differences in processing the inputs at sentence or text level make the classification too easy, we form two additional groups along this criterion, which are studied in a separate classification.
This also aligns with the fact that tools actually operate sentence-wise.
Including the group of all tools, this results in five different groups in our study.
We train the classifiers for the latter groups using our datasets:
the first training scenario uses our Twitter data, called corpus C1, by considering the tweets of all hashtags mentioned in \autoref{sec:data}.
The Wikipedia and the newspaper corpora of the languages are used to generate a second dataset (called C2).
Thirdly, we combine both datasets to form a third dataset (C3) for training.
We consider the Europarl corpus (C4) separately.
Using the development sets, we perform a parameter study on the size of the hidden layer ($5$--$300$, the optimizer (SGD and Adam), the learning rate ($0.001$--$0.1$) and the number of training epochs ($5$--$100$).
This brings the total number of experiments 
to \numprint{432000}, that is \numprint{180} per model, for each of the \numprint{800} combinations described above for each of the three languages.
We double the number of experiments with the German corpus to compare the differences when using a scaler on the feature data; however, we find no significant differences when using the NN.
\autoref{tab:big_table_raw} in the appendix shows a subset of the $F_1$-scores of the development and test sets for English (based on scikit-learn); \autoref{tab:big_table_means} shows statistics of the distributions of all scores. %

\begin{figure}
    \centering
	\includegraphics[width=.5\linewidth,trim={3cm 0.75cm 5cm 2cm},clip]{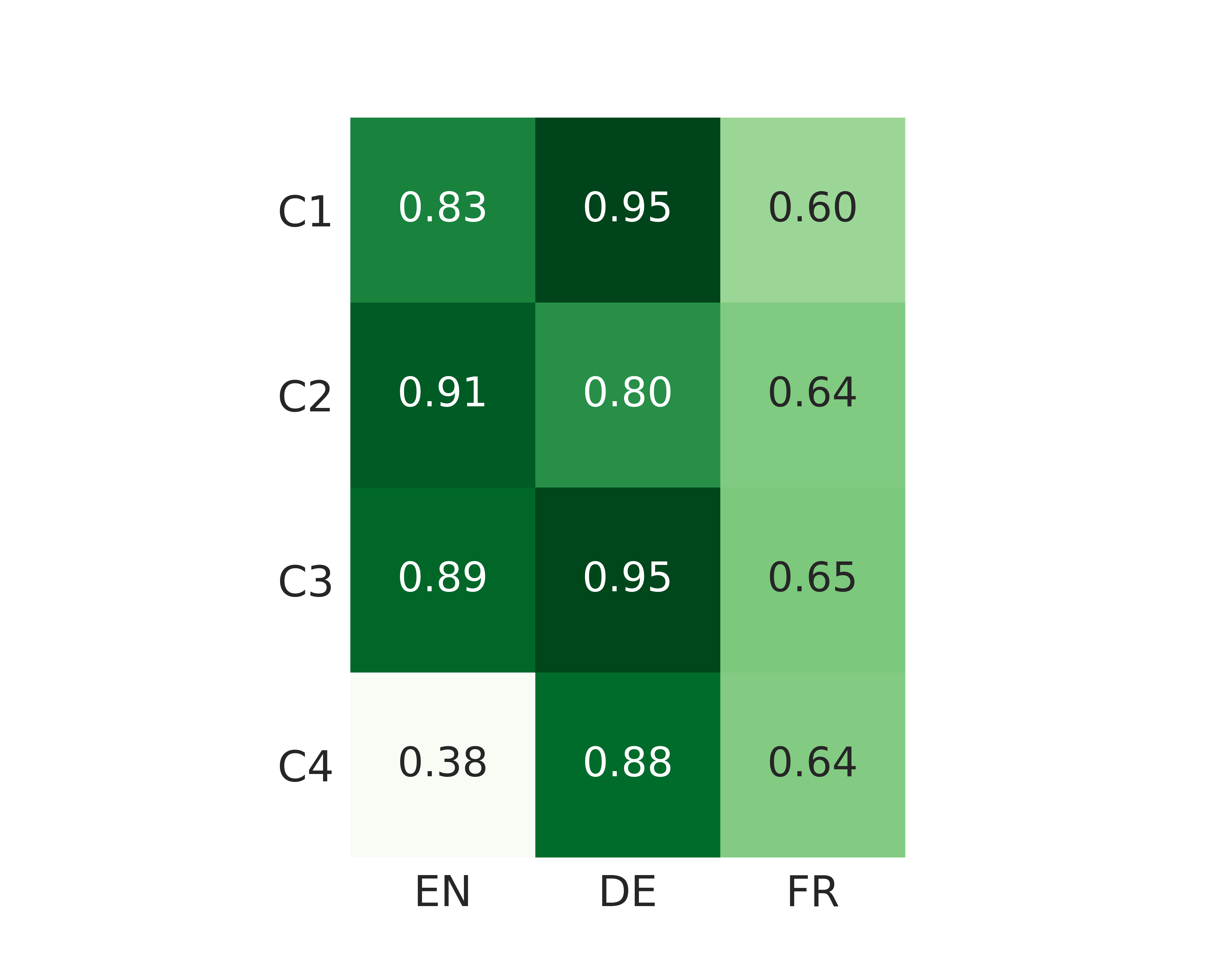}
    \captionsetup{skip=5pt}
    \caption{Results of the NN-based classifier: mean $F_1$-scores over all languages and corpora, averaged over all chunk sizes and then over all normalization variants.}
    \label{fig:results_mean_heatmap}
\end{figure}

If all tools predicted the same sentiment distributions for the same input texts, the trained classifier would not be able to distinguish these tools based on the statistical moments of these distributions.
However, we detect high $F_1$-scores across almost all experiments, combinations, and languages, as shown by the mean $F_1$-scores in \autoref{fig:results_mean_heatmap}.
\begin{description}[style=unboxed,leftmargin=0cm]
    \item[English] Considering corpus C3 without normalization of the sentiment values, we reach an $F_1$-score of $0.927$ averaging over the results using different chunk sizes, with the smallest $F_1$-score of $0.867$ at a chunk size of \numprint{50} and the largest of $0.988$ at a chunk size of \numprint{1000}.
    Normalizing the sentiment values according to $N_3$ lowers performance to a still high $F_1$-score of $0.797$, mainly due to effects of small chunk size (i.e., $0.684$ at size \numprint{50}; at size \numprint{1000} the $F_1$-score reaches $0.927$).
    The other normalization methods, which leave more room for different sentiment values, result in higher average scores of $0.824$, $0.903$, and $0.918$.
    Looking closer at different corpora, we discover an average $F_1$-score of $0.86$ on Twitter (max $0.959$) and $0.959$ on Wikipedia and the NYT (here even reaching $1.0$).
    This suggests that the tweets are more difficult to process.
    In the case of grouping tools according to whether they produce discrete or continuous sentiment values, we obtain mostly similar results, with higher scores for the latter group. We attribute this to the fact that they leave more room for different sentiments. 
    If we consider the group that emerges when sentiment values are generated from averaging over sentences, we see a decrease from $0.911$ to $0.883$ for tools with discrete sentiment values and an improvement from $0.978$ to $0.992$ for those with continuous values.
    This is expectable, since in the case of discrete sentiments we only perform a binary classification of more difficult-to-separate data.
    These results support our claim that one can predict sentiment analysis tools based on simple statistics of the value distributions they produce, regardless of the text genre and the underlying method.
    Our results on German and French show that this is predominantly true for these languages as well:
    \item[German] As with the English corpora, we see a quite high average $F_1$-score of $0.982$ on C3, score of $0.985$ on Twitter and of $0.933$ on C2.
    The classifier works perfectly with discrete sentiments on C3 and on Twitter data ($F_1$-score: $1.0$) and a score close to $0.997$ in the case of C2.
    Results on the sentence only data is equally high.
    Scores on continuous data is a bit lower, with the sentence-based experiments performing on the same level again.
    \item[French]
    The values for the French corpora are generally lower, with an average of $0.694$ for C3, $0.648$ for Twitter, and $0.683$ for Wikipedia and Newscrawl.
    We find that the classifier has difficulties separating the tools that provide discrete sentiment scores ($F_1$ score: $0.415$).
    This can be seen in the boxplots in appendix \autoref{fig:boxplot_defrwiki_sputnikv}, where the BERT-based models produce very similar statistical distributions.
    However, the tools producing continuous sentiment values are well separated (average $F_1$-score: $0.996$).
    \item[Europarl] While we obtain very good results on the Twitter, Wikipedia, and news corpora, our classifier is not able to distinguish the tools based on the sentiments they generate for the English Europarl corpus.
    The $F_1$ value drops to $0.169$ for most configurations and groups, except for the continuous data based on sentences, where we reach a value of $0.967$.
    However, the values for the German and French corpora are on par with those for the other corpora ($0.954$ and $0.696$, respectively).
    This implies indistinguishable distributions of sentiment values in the Europarl corpus, at least when considering English sentiment tools, which we have previously shown to be distinguishable in the other corpora.
    Given the high scores for the German Europarl corpus, we can probably rule out the possibility that this result is due to the simplicity and brevity of the Europarl texts; rather, it is likely a result of too little training data.
\end{description}
\subsection{Other classifiers}
\label{sec:Other_classifier}
We alternatively experimented with \textit{Support Vector Machine}s (SVM) and \textit{Decision Tree}s (DT) to see if they lead to similar results.
More specifically, we trained $k$-nearest neighbors (KNN, k=5), SVM with linear and rbf kernels, and a DT with a maximum depth of 5.
The best results were obtained with KNN with an average $F_1$-score of $0.903$ on C3 (with $0.981$ max).
Tools operating on Twitter are also well separated with an average score of $0.845$; the C2 corpus yields an $F_1$-score of $0.94$.
Consistent with the results based on the NN classifier, Europarl is problematic with a low average score of $0.213$.
The other tools perform on average on C3 as follows:
DT: $0.551$ ($0.594$ max), SVM-linear: $0.831$ ($0.881$ max), SVM-rbf: $0.811$ ($0.843$ max).
On the German corpora, the KNN and SVM classifier perform extremely well, often reaching $F_1$-scores of or close to $1.0$, over all configurations and groups analyzed.
In line with our experiments using a NN, the scores on the French corpora are lower overall, due to the results on the discrete sentiment values.
On the C3 corpus, however, the SVM-linear still achieves an average $F_1$-score of $0.717$ ($0.831$ max).
The appendix shows visualizations of the DTs trained on C3 (\autoref{fig:decision_trees_all}) and the Europarl corpus (\autoref{fig:decision_trees_eu}), each for EN, DE and FR.
\subsection{Monte Carlo sampling}
\label{sec:monto_carlo}
To investigate, whether the results obtained above hold under different sampling regimes, we perform a Monte Carlo simulation, where we draw with replacement a 1\%-sized subset of the total sentiment predictions.
For each tool we generate $m$ such samples and calculate sample statistics, that is, two variants of entropy and the other moments (see \autoref{sec:nn}), for each.
That way we produce $n\times m$ feature vectors of length $l$, where $n$ is the number of tools analyzed in the experiment, $m\in \{10, 50, 100, 500, 1000\}$, and $l\in [1, 15]$ is the number of features.
Subsequently, we train an SVM by means of scikit-learn, that predicts a tool based on the generated feature vectors.
Ideally, if the tools produce similar predictions, the SVM would be unable to find a separating hyperplane, resulting in low $F_1$-scores.
However, the $F_1$-scores are significantly higher than expected, see \autoref{fig:svm_res_all} (and \autoref{fig:svm_res_all_de_fr_and_ep} in the appendix for results on DE and FR) for the C3 dataset, reaching $1.0$ for German, $\approx 1.0$ for English and $\approx 0.86$ for French.
These results agree with low distance correlation scores of \autoref{fig:dcor_all} and the obtained results so far.
\begin{figure*}
    \centering
	\includegraphics[width=0.44\linewidth]{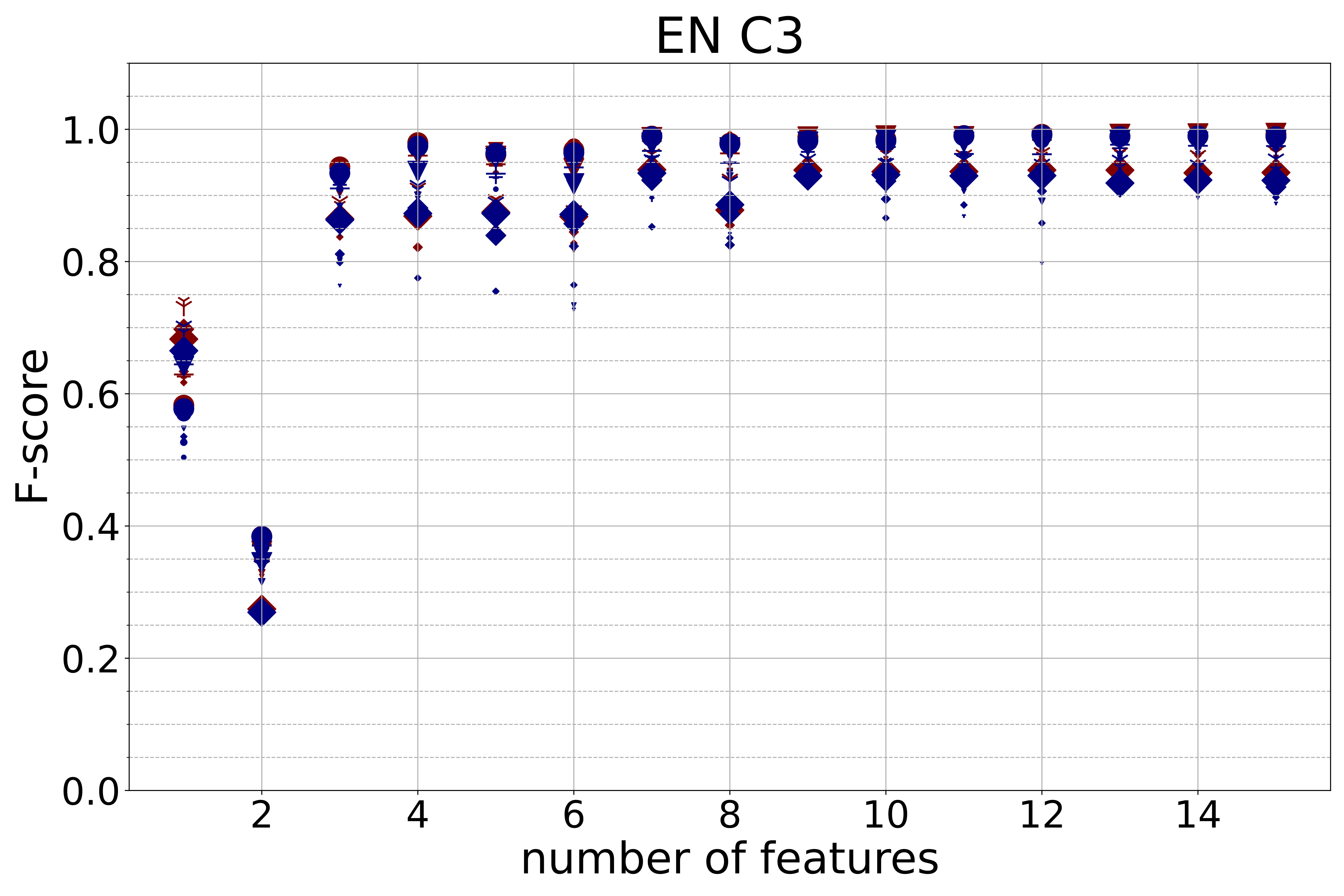}
	\includegraphics[width=0.44\linewidth]{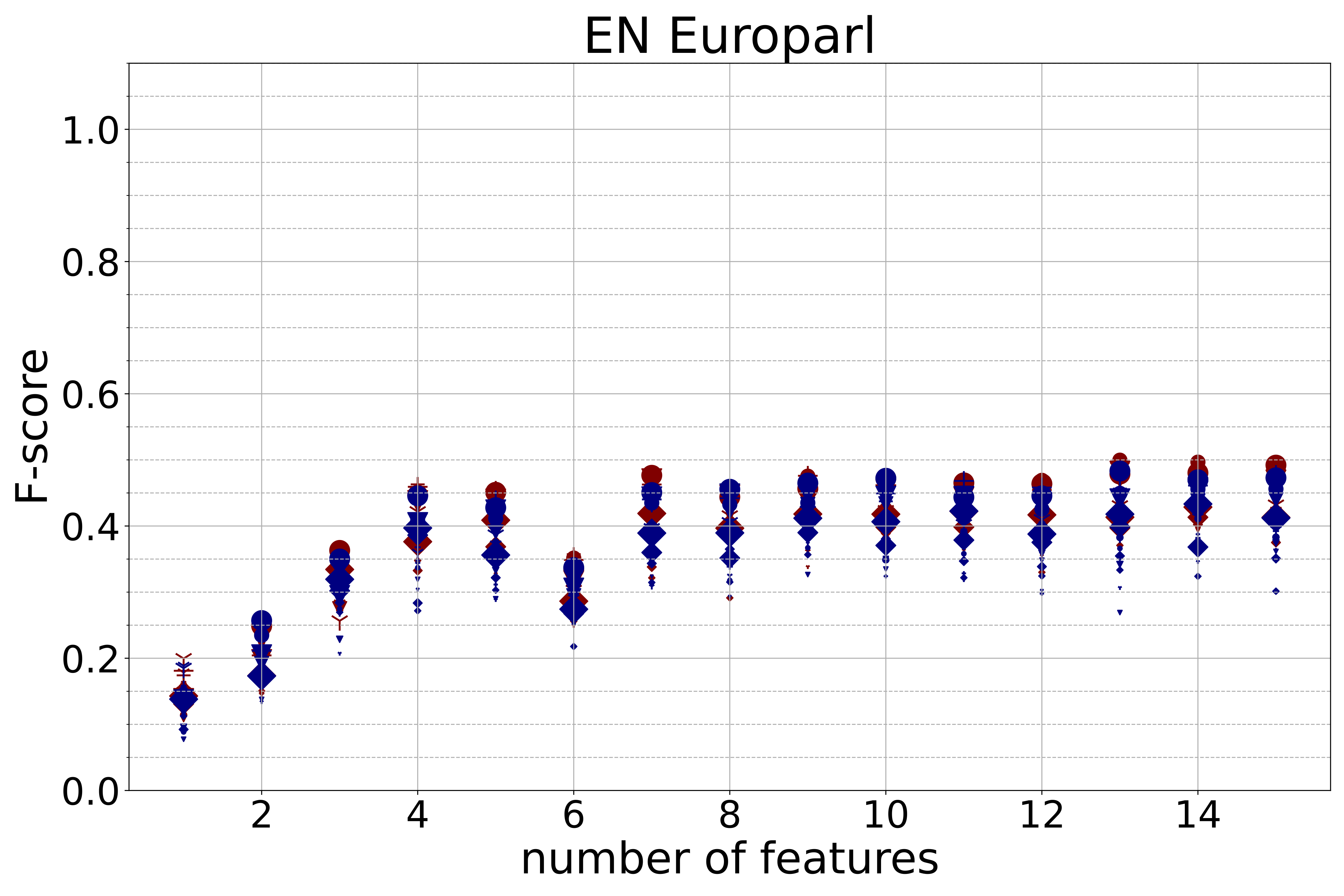}
	\includegraphics[width=0.1\linewidth, raise=1.2cm]{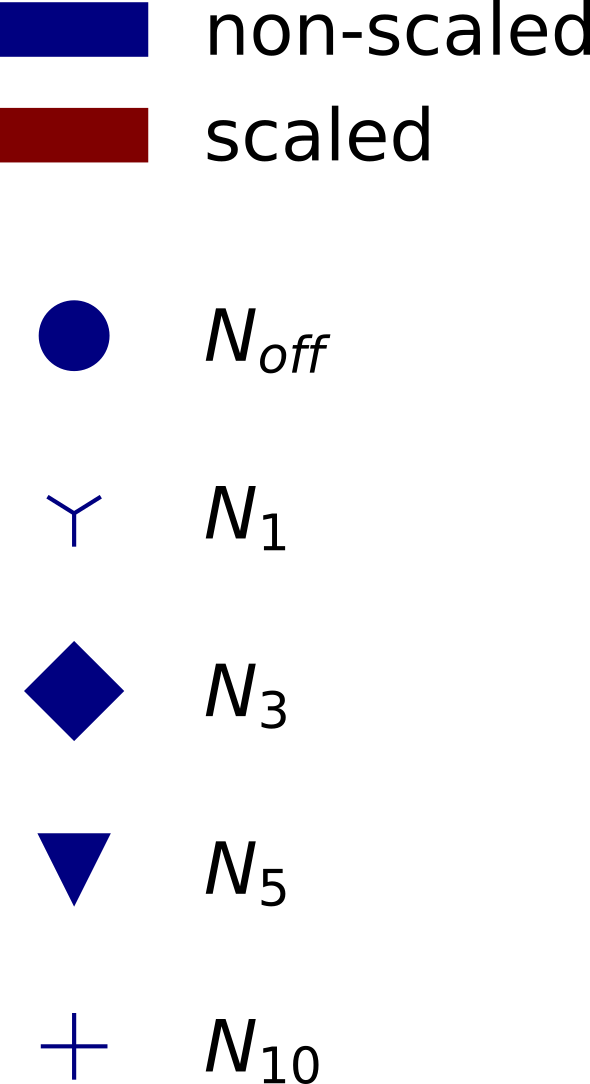}
    \captionsetup{skip=5pt}
    \caption{$F_1$-scores of the SVM classifier for the C3 and Europarl corpora. The x-axis indicates the number of features, which are selected randomly for each feature vector size independently.}
    \label{fig:svm_res_all}
\end{figure*}
As in the previous sections, the only experiment, where the SVM struggles, is the English Europarl corpus.
\section{Discussion}
\label{sec:discussion}
Using microblogging data and data from three other genres, our results show that we are able to distinguish sentiment analysis tools based on simple statistical features of their sentiment value distributions.
Of course, this result depends on the tools we choose and the corpora we use for evaluation.
In particular, the size of the corpora and the subsequently generated training and test samples have influence on our findings.
However, looking at the mean $F$-scores averaged over the different normalization methods and sample sizes, we see that our hypothesis is not falsified.
That is, different sentiment analysis tools produce different sentiment distributions that can be distinguished \textit{and} identified after the fact.
In other words, \textit{you shall know a sentiment tool by the traces it leaves}. 
In light of our introduction, this means that the increasing use of NLP tools to support empirical claims in, e.g., the social sciences, education, or the humanities, faces the difficulty of substantiating the validity of its findings.
At the very least, these approaches should clarify the extent to which their results are algorithmically biased by the chosen tool and how those results change by using other tools.

We selected Wikipedia and news corpora to contrast the short, rather subjective microblogging texts with texts from other genres and to test whether our hypothesis is also confirmed by them. 
The use of Europarl data served the comparison under the condition of considering texts with the same meaning for the targeted languages.
The differences in the results for Europarl suggest that additional languages should be investigated.
We considered a number of tools to cover different aspects of sentiment detection methods, i.e., we evaluated lexicon-based and model-based approaches.
Since the use of SOTA transformer-based models has been shown to be very effective in various NLP tasks, we have also included them.
We show that these tools produce partially similar results. This could be due to the fact that they generate discrete sentiment values or use the same model architecture.
To control for the first of these candidates, we used different normalization variants to discretize the results of all tools. We found that classifiers trained with normalizations that partition their range of values to produce more discrete values have a higher $F_1$-score than classifiers based on normalizations that result in less discrete sentiment values.
To further substantiate this research, more tools for more genres should be considered.
\section{Conclusion}
\label{sec:conclusion}
We examined for three languages, with multiple corpora each, whether different sentiment tools are distinguishable with regards to their outputs.
The sentiment tools analyzed mostly use lexicons or transformer-based models trained with data from Twitter or reviews.
In several experiments, we trained classifiers that could identify a tool based on statistical moments of the sentiment value distribution it produced. 
This research shows that empirical studies relying on a single sentiment analysis tool may be algorithmically biased by the choice of that tool: other tools are likely to produce different, readily identifiable results for the same data.
To support this finding, we argue for an expansion of research on algorithmic biasing.
In future work, we plan to extend our experiments to more tools and corpora from other genres and languages.
This will include experimenting with pre-processing, sampling, and feature selection methods.

\newpage

\bibliography{anthology,custom}
\bibliographystyle{acl_natbib}

\newpage

\appendix

\section{Appendix}
\label{sec:appendix}
\begin{figure*}
    \centering
    \includegraphics[width=.49\linewidth]{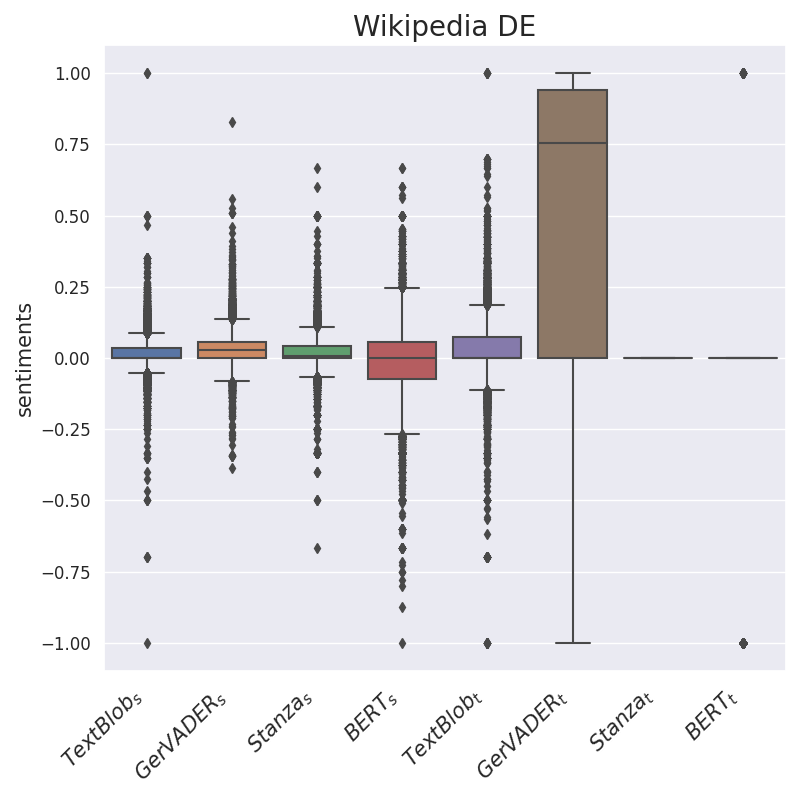}
    \includegraphics[width=.49\linewidth]{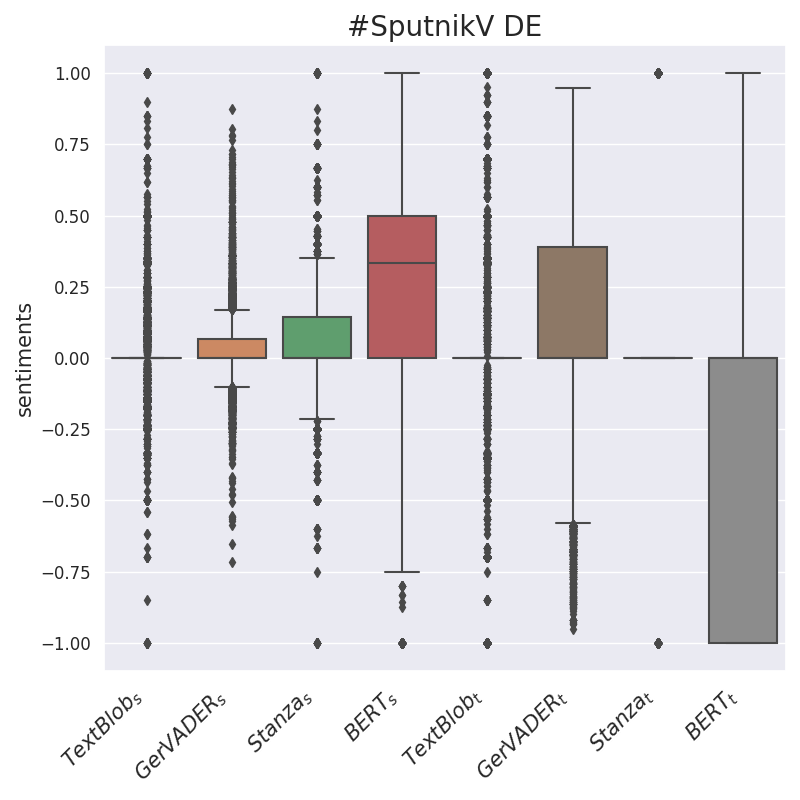}
    \includegraphics[width=.49\linewidth]{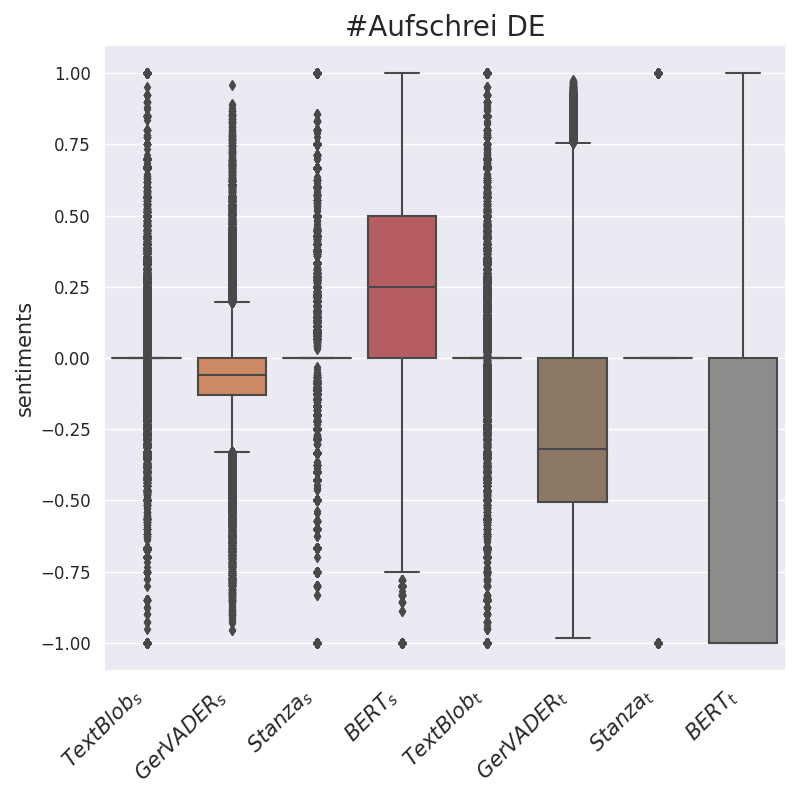}
    \includegraphics[width=.49\linewidth]{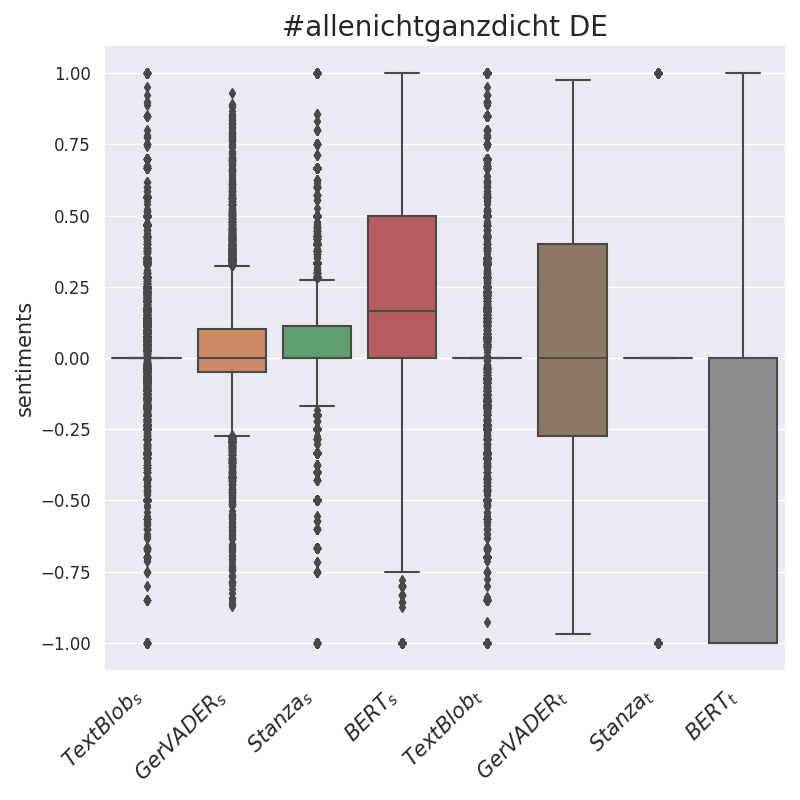}
    \includegraphics[width=.49\linewidth]{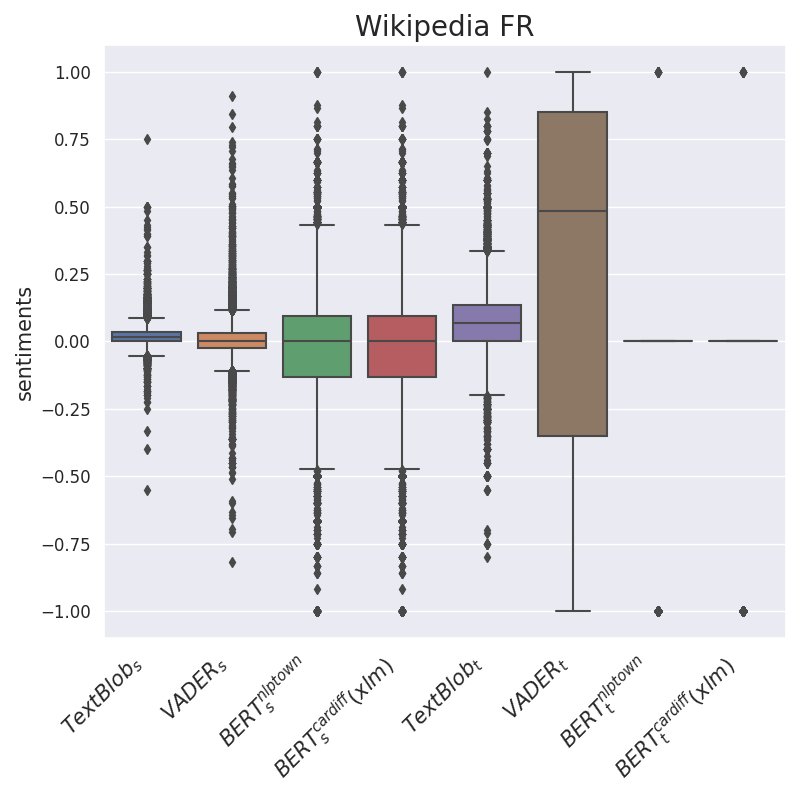}
    \includegraphics[width=.49\linewidth]{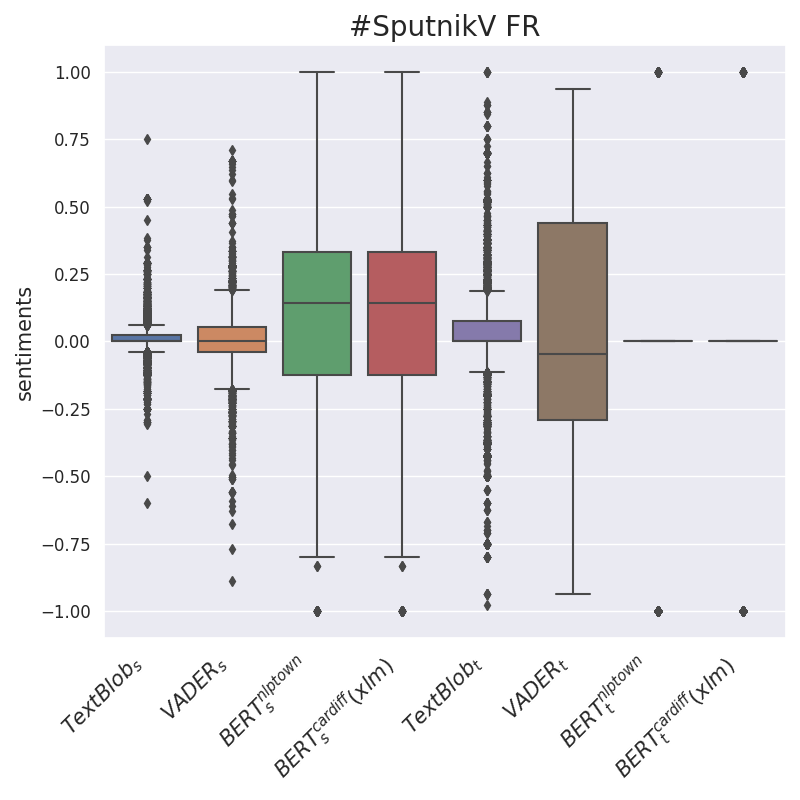}
    \caption{Sentiments of Wikipedia and Twitter hashtags \#SputnikV, \#Aufschrei and \#allenichtganzdicht for German and French language.}
    \label{fig:boxplot_defrwiki_sputnikv}
\end{figure*}

\begin{figure*}
    \centering
    \includegraphics[width=.49\linewidth]{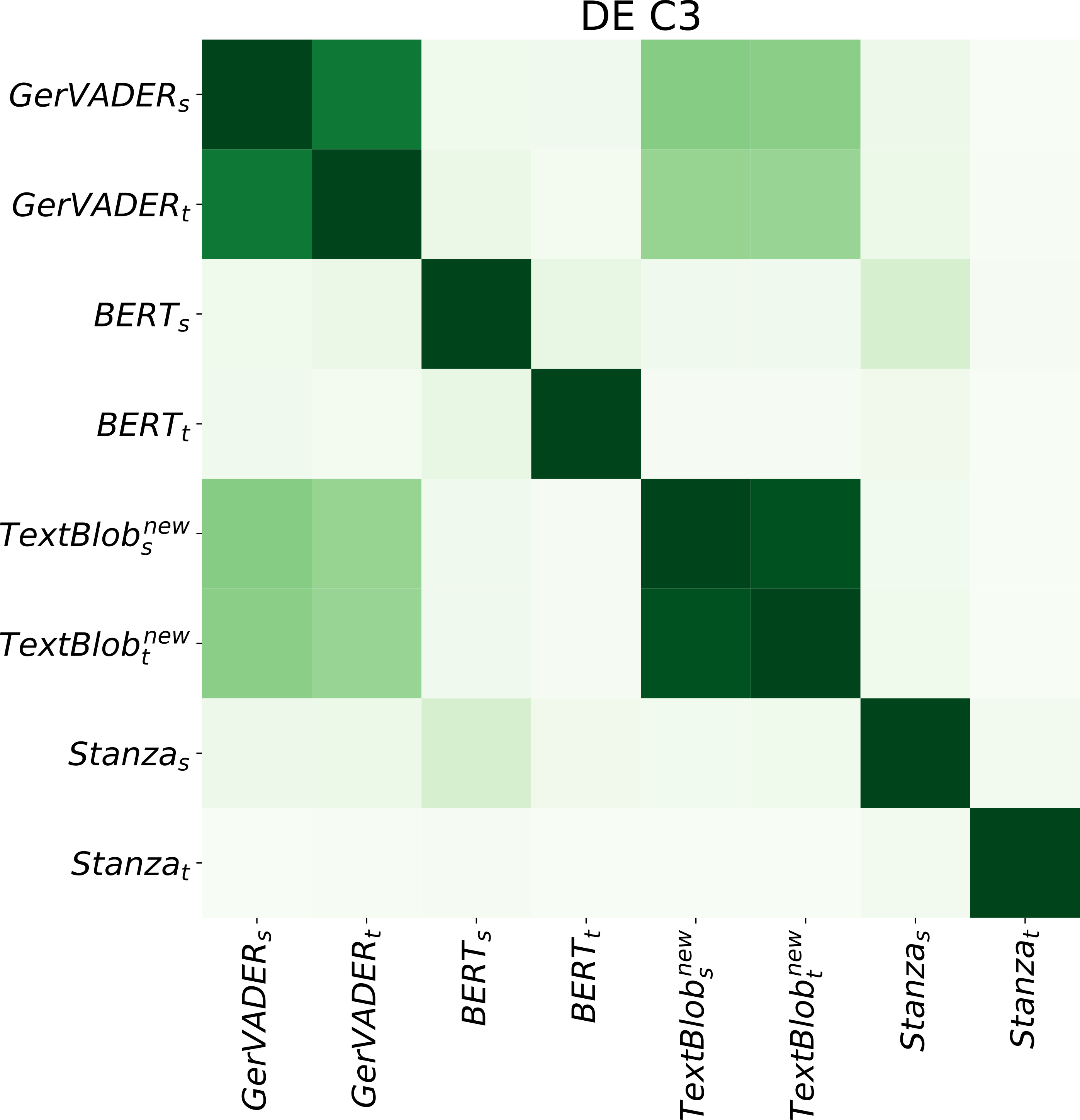}
    \includegraphics[width=.49\linewidth]{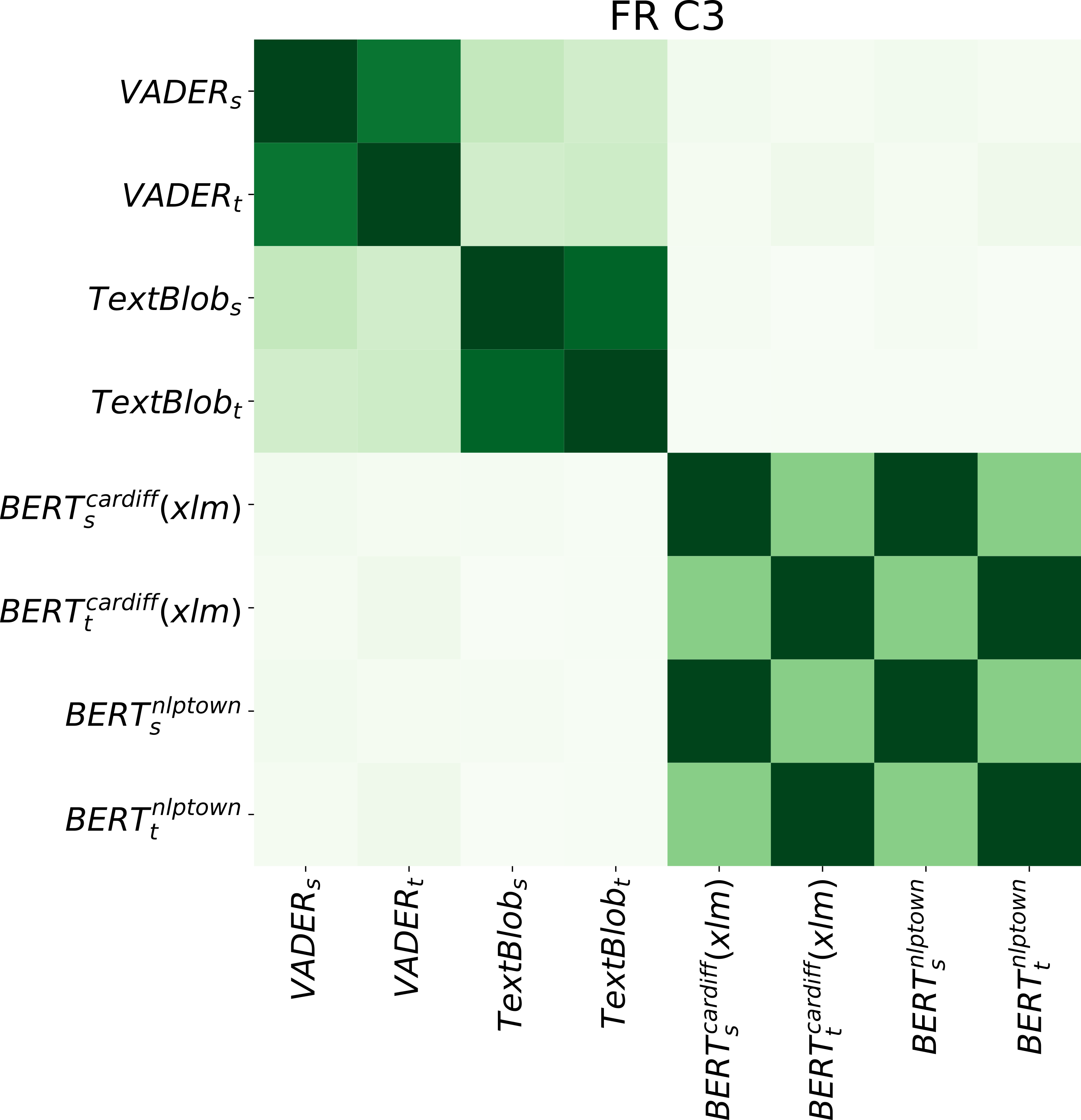}
    \includegraphics[width=.32\linewidth]{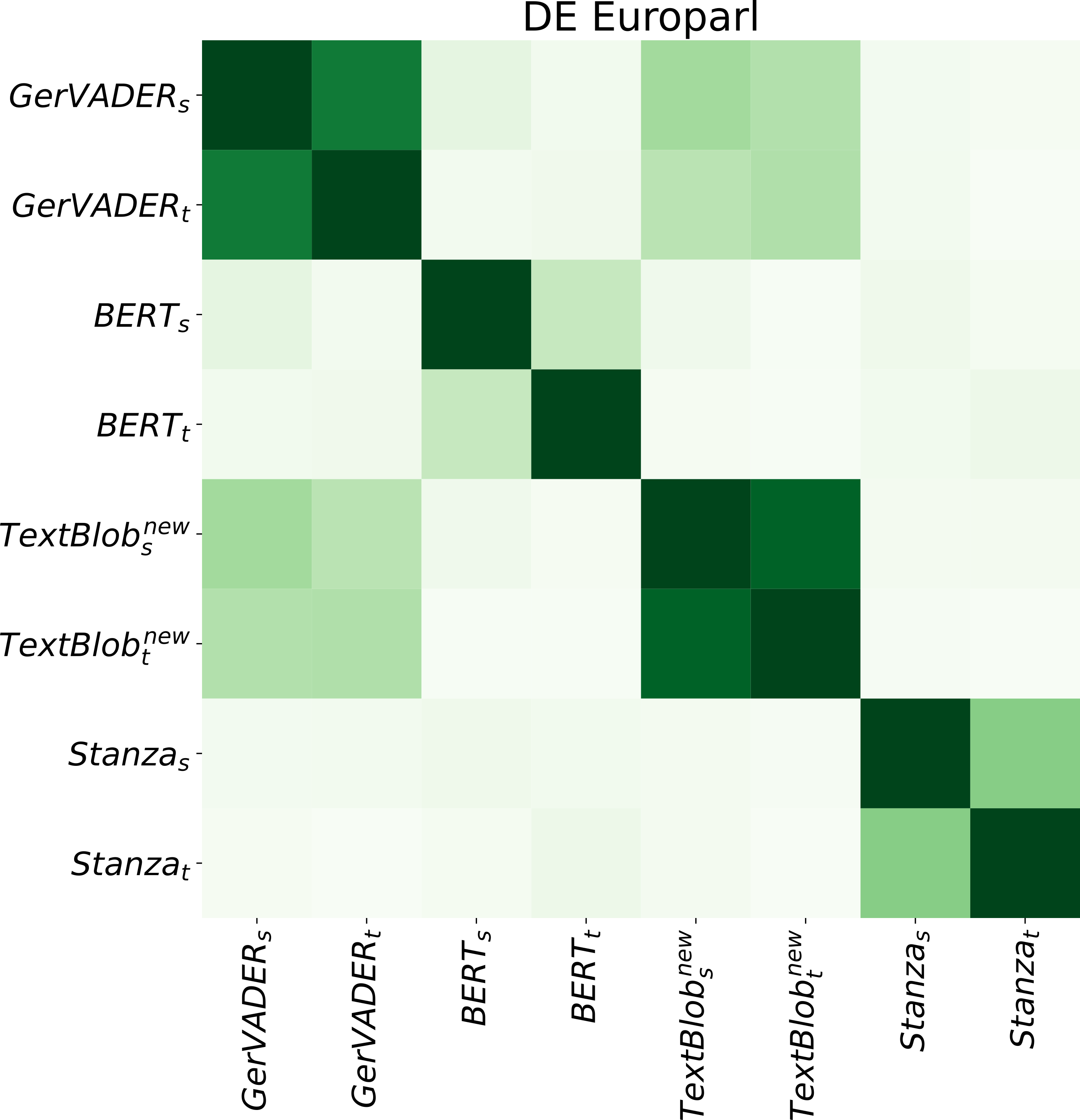}
    \includegraphics[width=.32\linewidth]{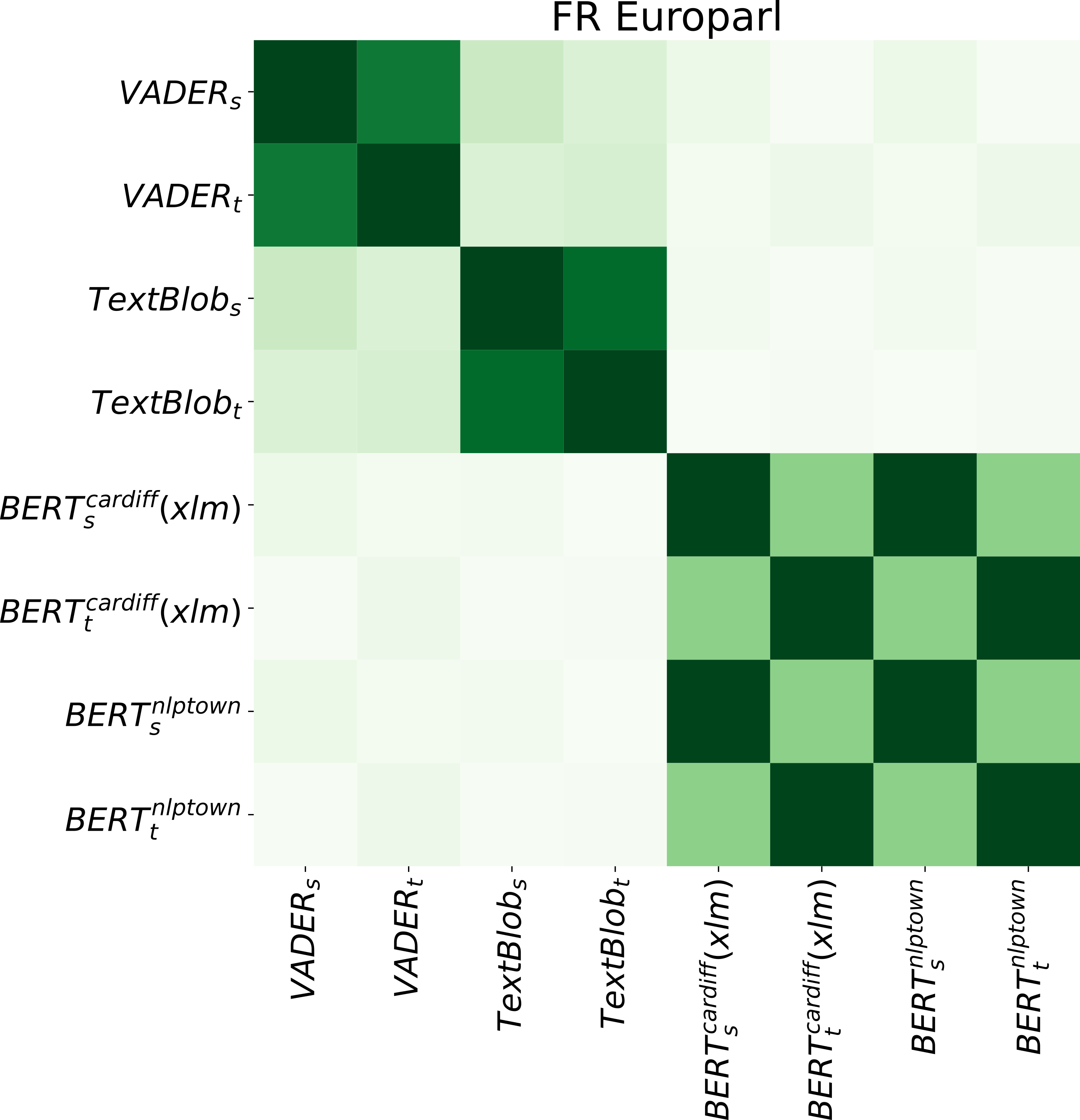}
    \includegraphics[width=.32\linewidth]{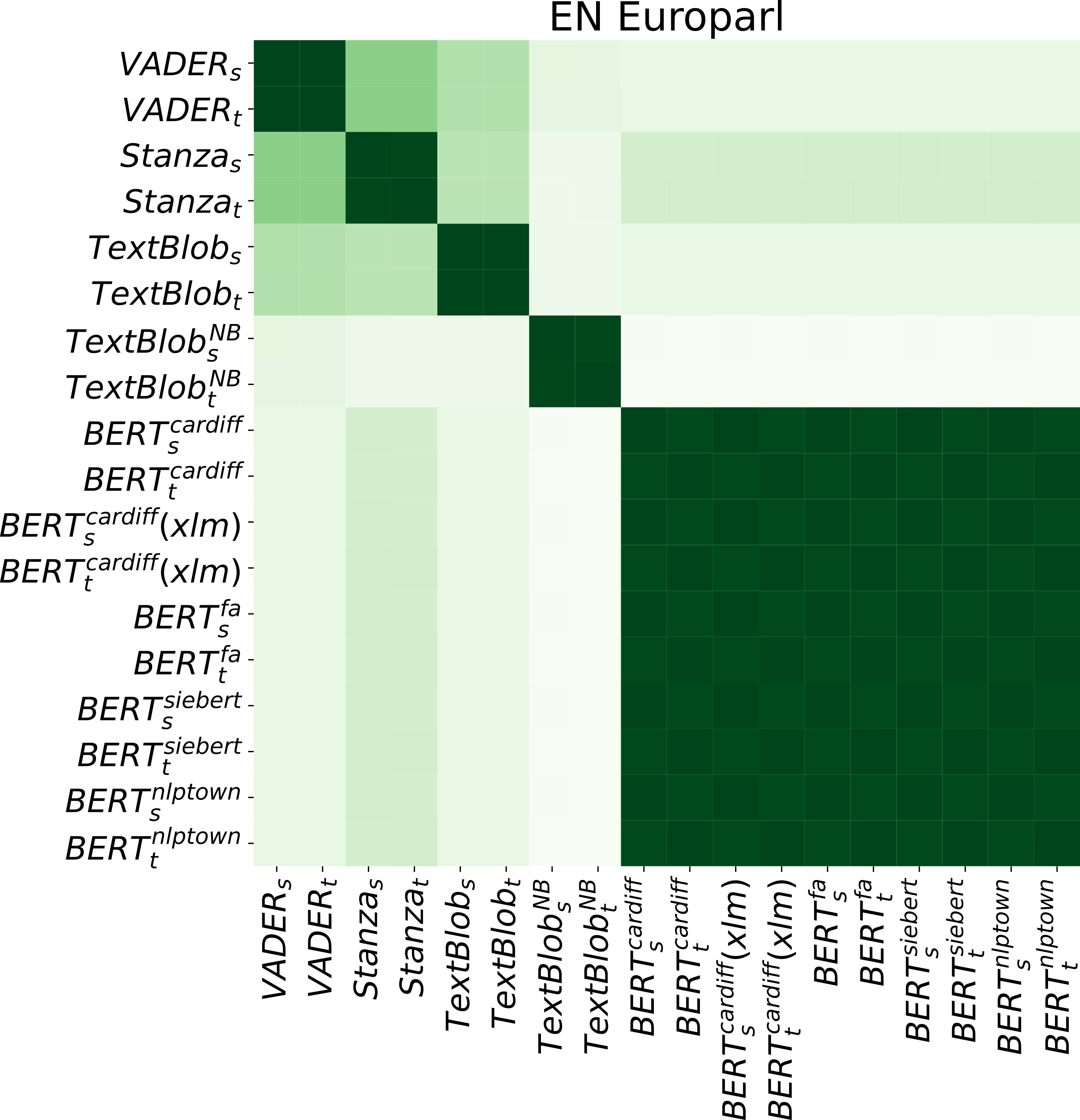}
    \caption{Distance correlation of sentiment scores for different tools on the C3 and Europarl corpora, with darker color indicating a higher correlation.}
    \label{fig:dcor_all_de_fr_en}
\end{figure*}

\begin{figure*}
    \centering
    \includegraphics[width=.49\linewidth]{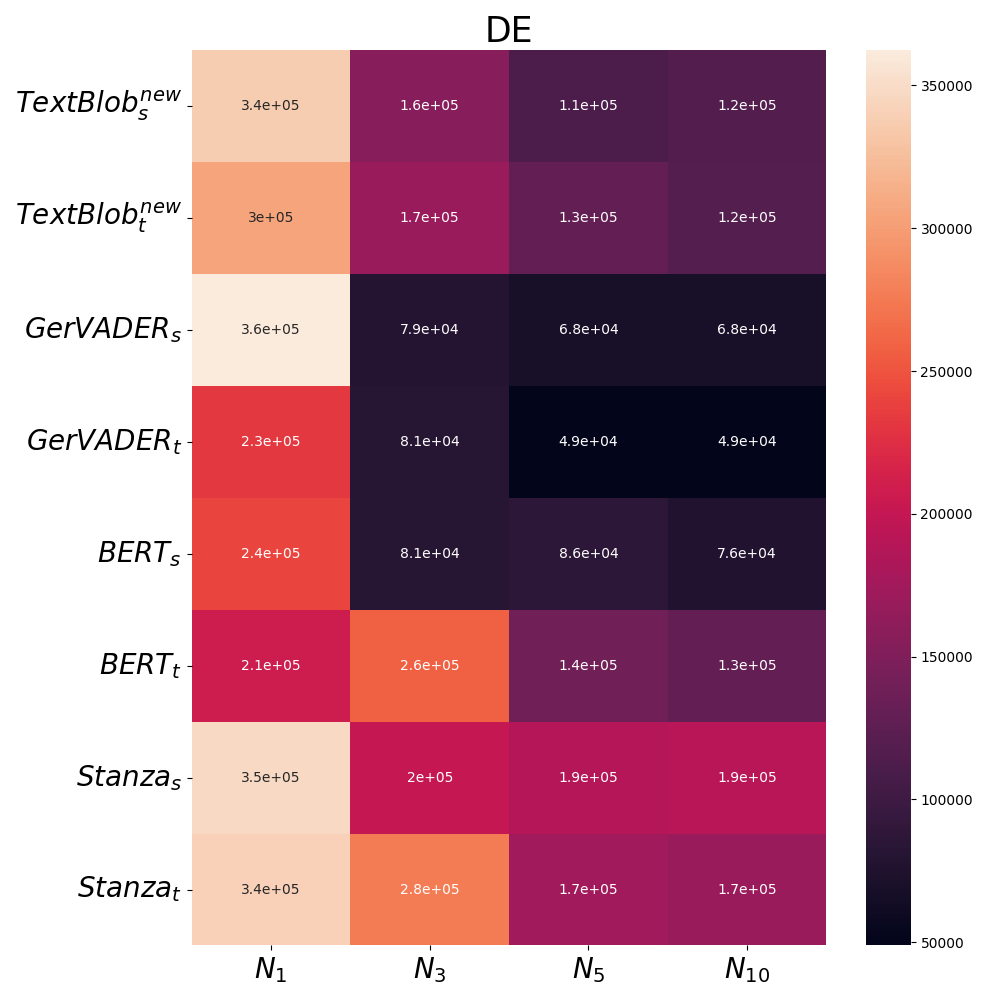}
    \includegraphics[width=.49\linewidth]{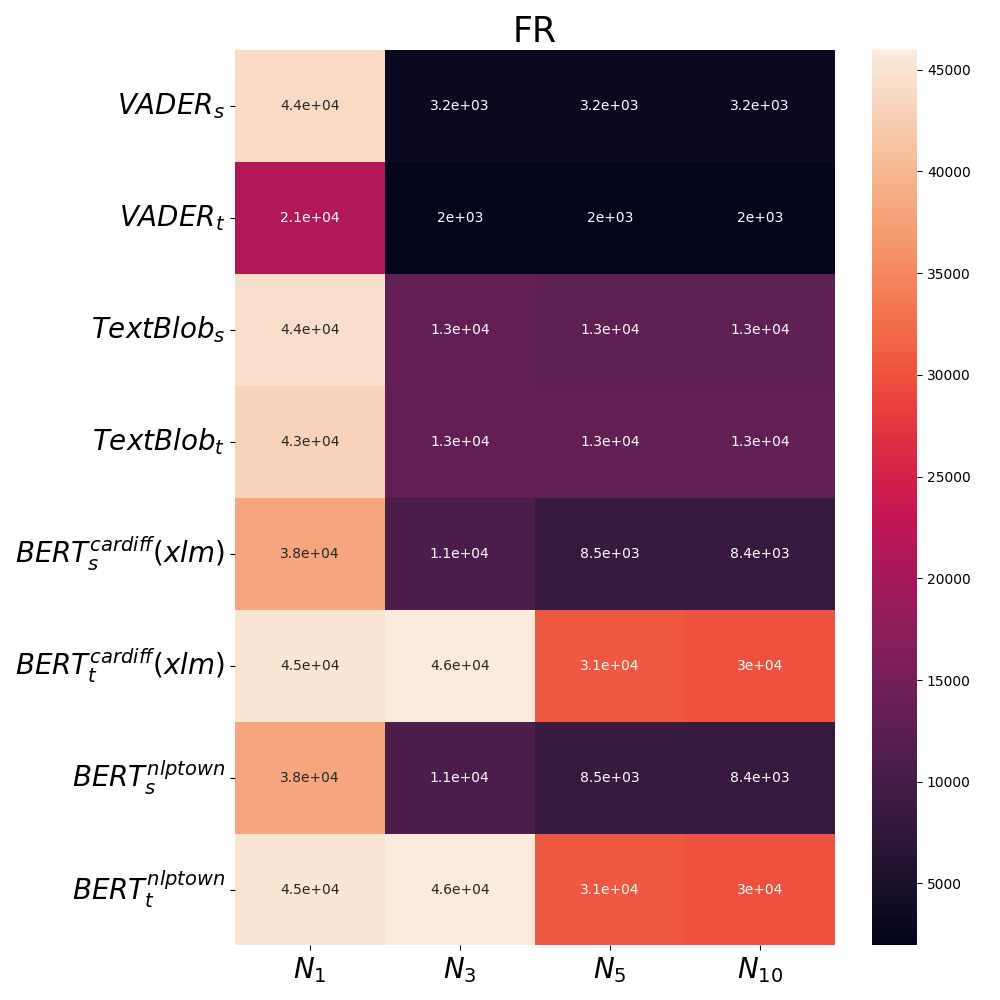}
    \caption{Agreement counts to the majority vote of the sentiment tools for German and French C3 corpus.}
    \label{fig:voting_de_fr}
\end{figure*}

\begin{figure*}
    \centering
    \begin{tabular}{ccc}
    \includegraphics[width=0.44\linewidth]{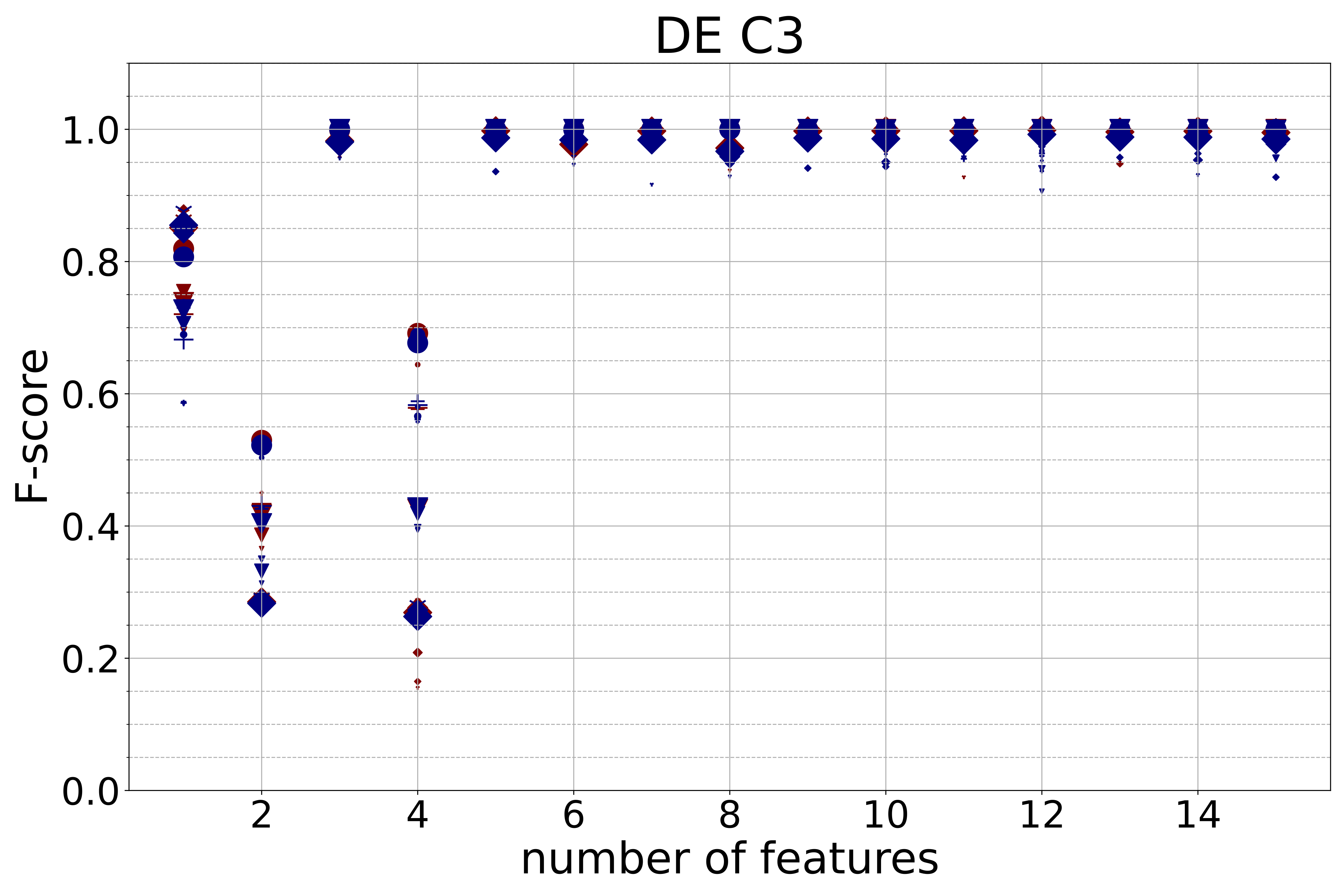} &
	\includegraphics[width=0.44\linewidth]{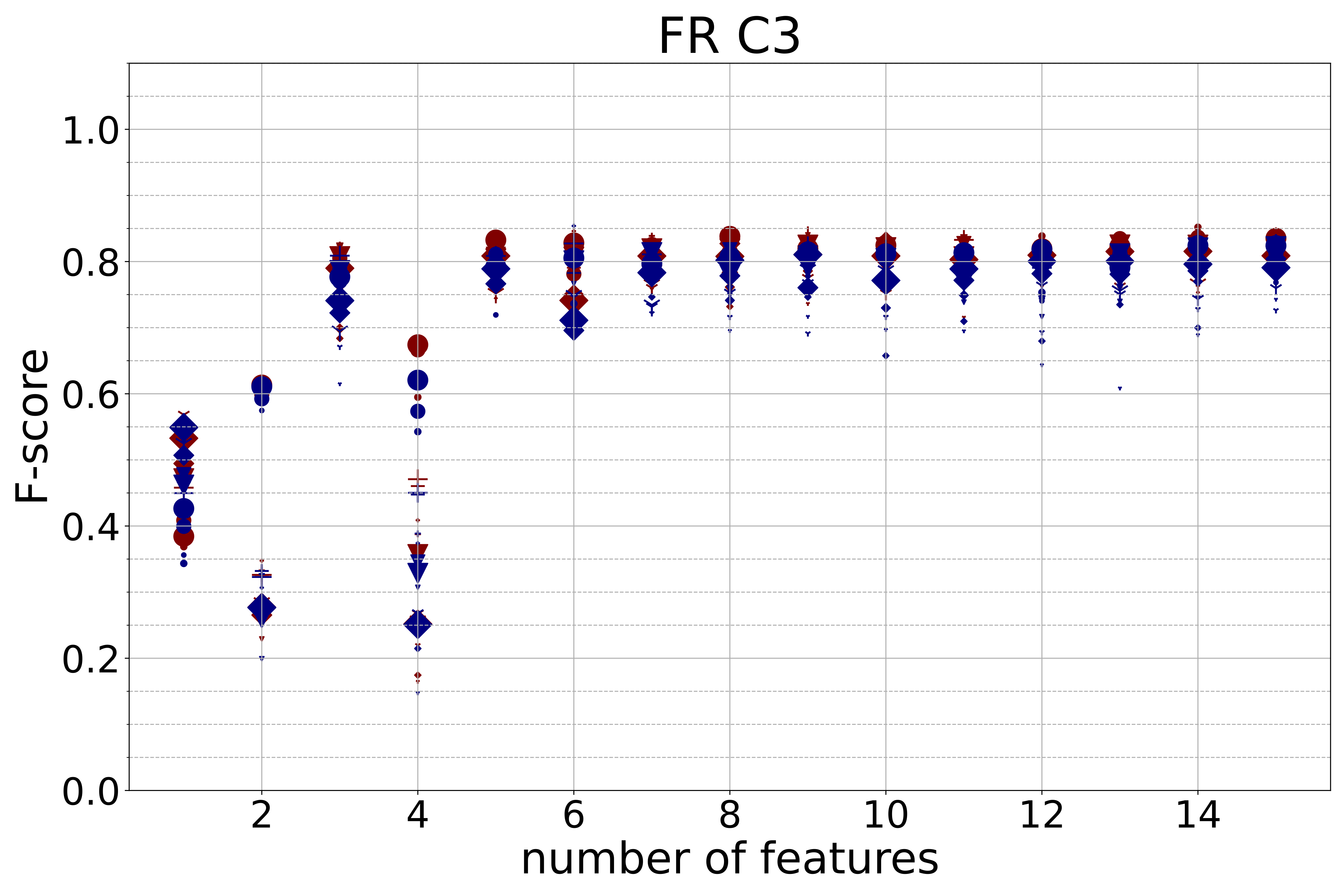} &
	\includegraphics[width=0.1\linewidth]{img_svm_results_legend.png}\\
    \includegraphics[width=0.44\linewidth]{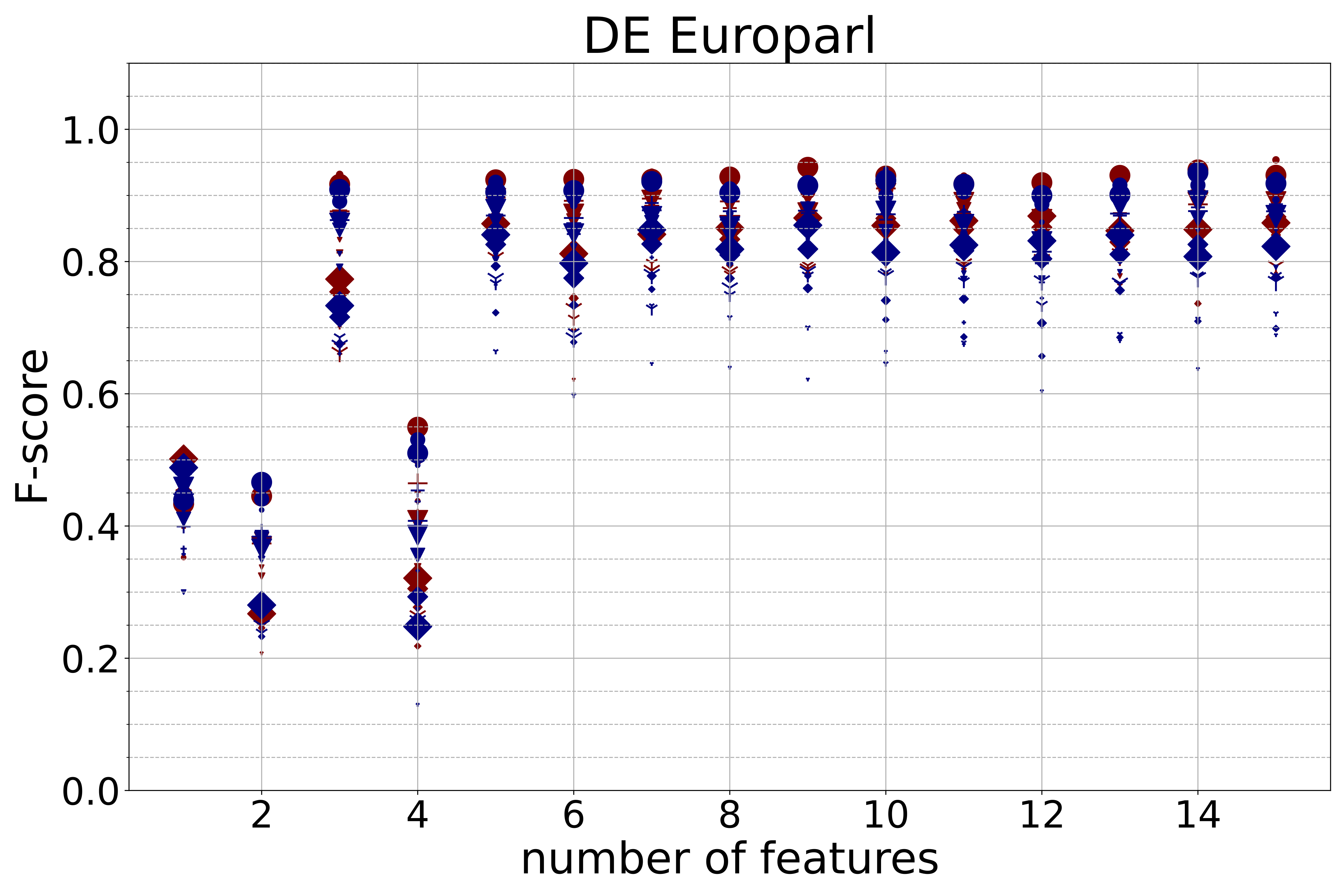}&
	\includegraphics[width=0.44\linewidth]{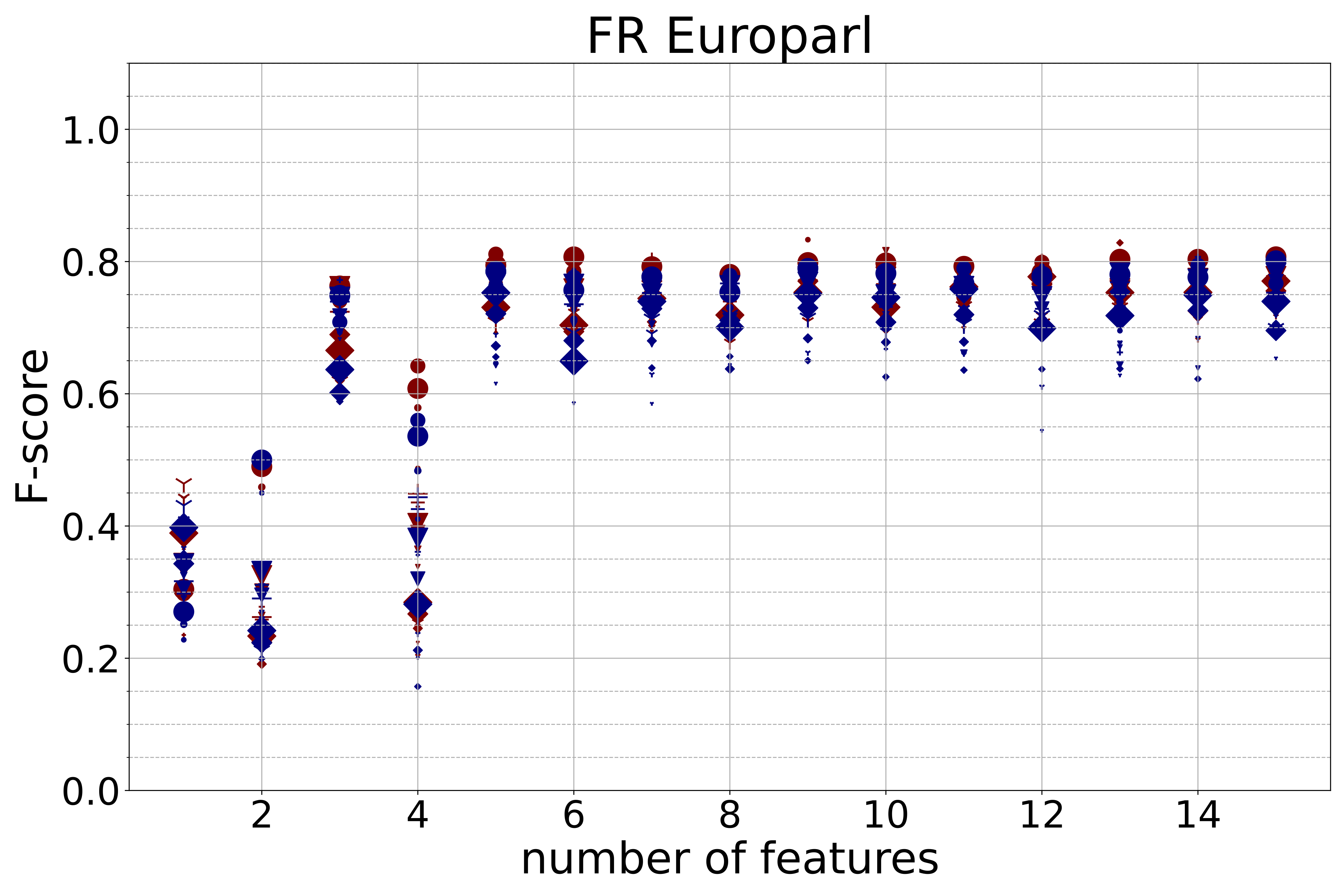}&
	\end{tabular}
    \caption{$F_1$-scores of the SVM classifier for the C3 and Europarl corpora. The x-axis indicates the number of features, which are selected randomly for each feature vector size independently.}
    \label{fig:svm_res_all_de_fr_and_ep}
\end{figure*}

\begin{figure*}
    \centering
    \begin{subfigure}[b]{0.90\textwidth}
        \centering
        \includegraphics[width=.90\linewidth, trim={45.5cm 0 35cm 0},clip]{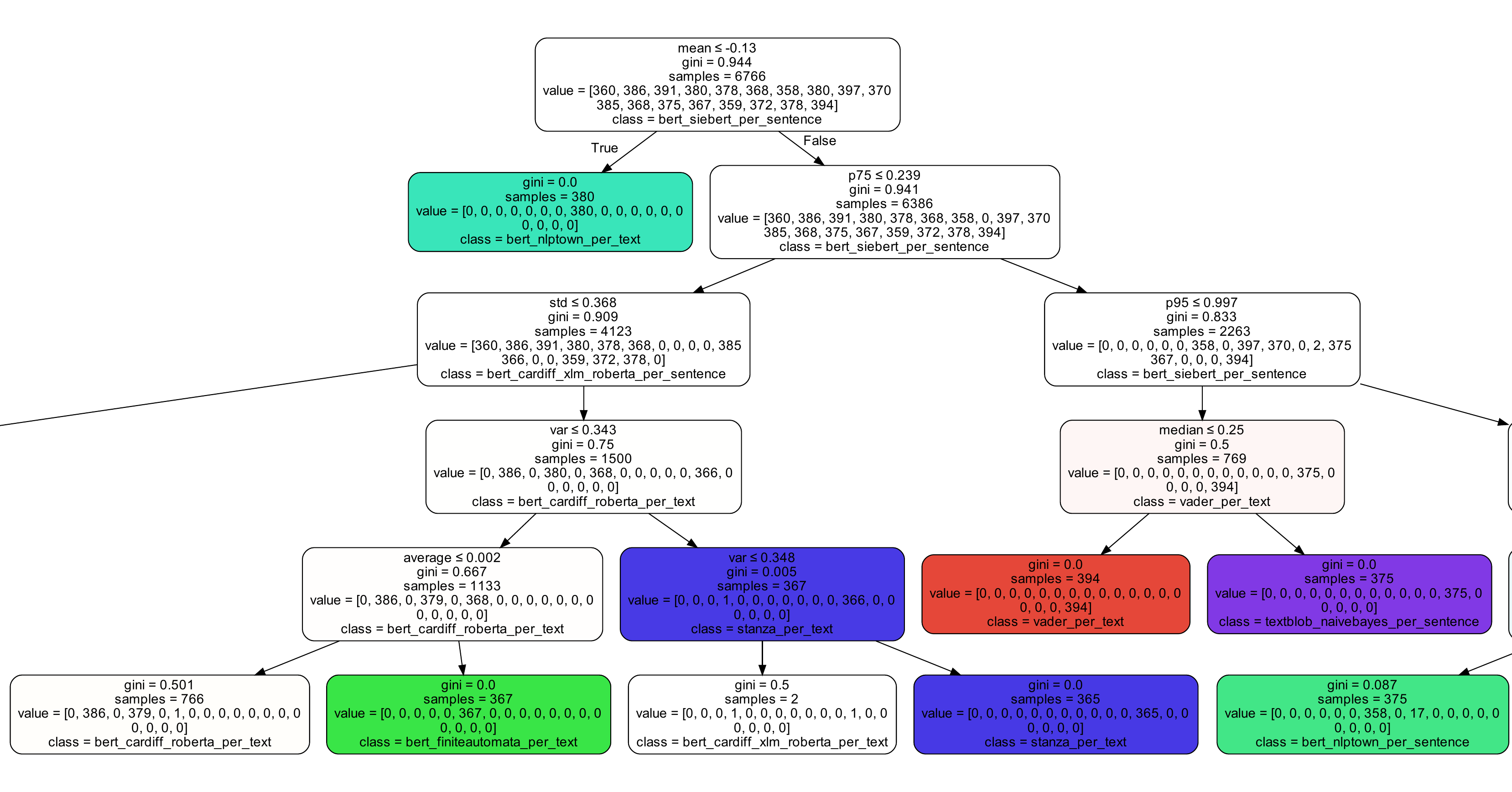}
        \vspace{-0.3cm}
        \caption{EN Decision Tree}
    \end{subfigure}
    \begin{subfigure}[b]{0.90\textwidth}  
        \centering 
        \includegraphics[width=.90\linewidth, trim={0 0 0 0},clip]{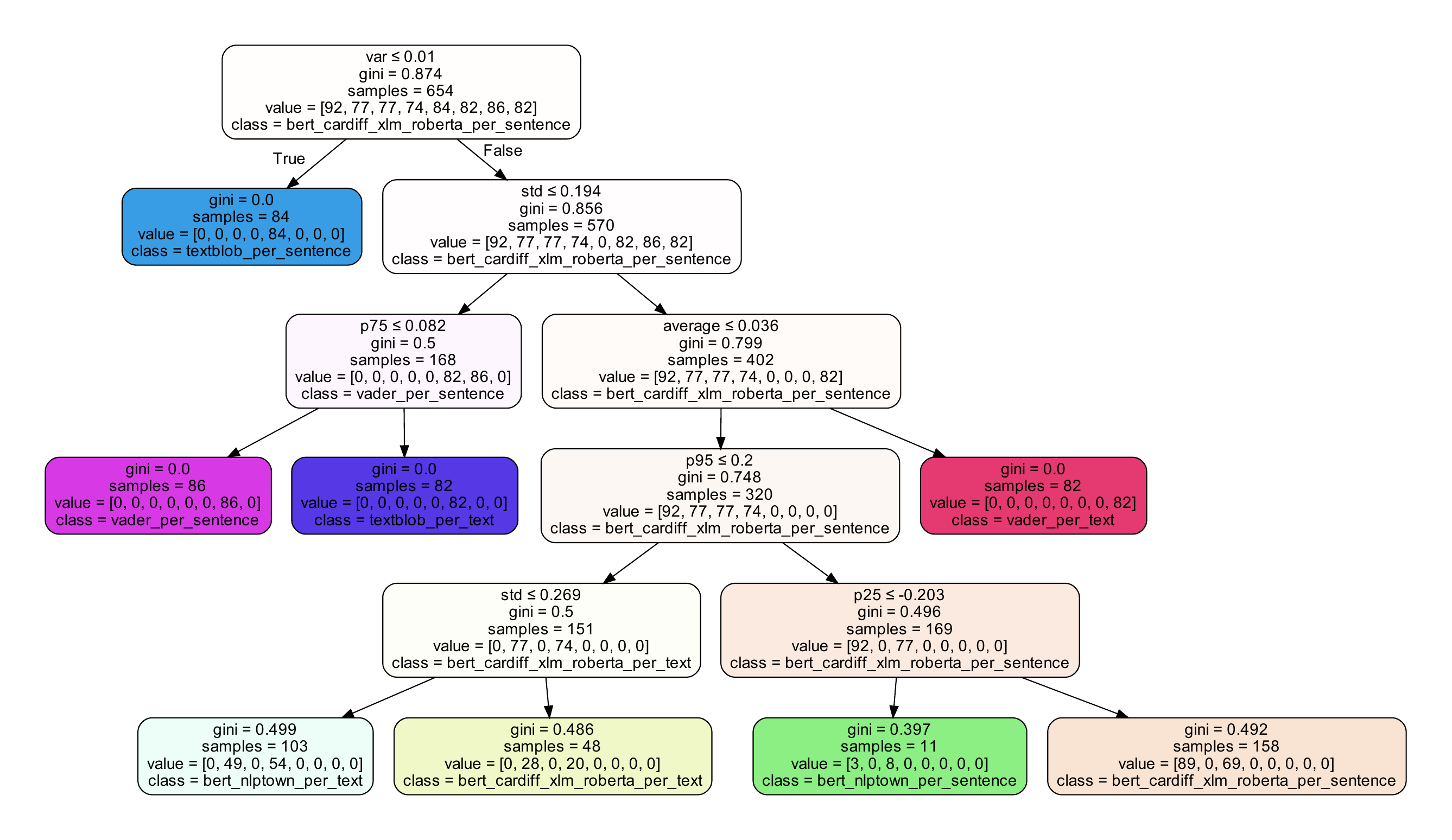}
        \vspace{-0.5cm}
        \caption{FR Decision Tree}
    \end{subfigure}
    \begin{subfigure}[b]{0.80\textwidth}   
        \centering 
        \includegraphics[width=.90\linewidth, trim={0 0 0 0},clip]{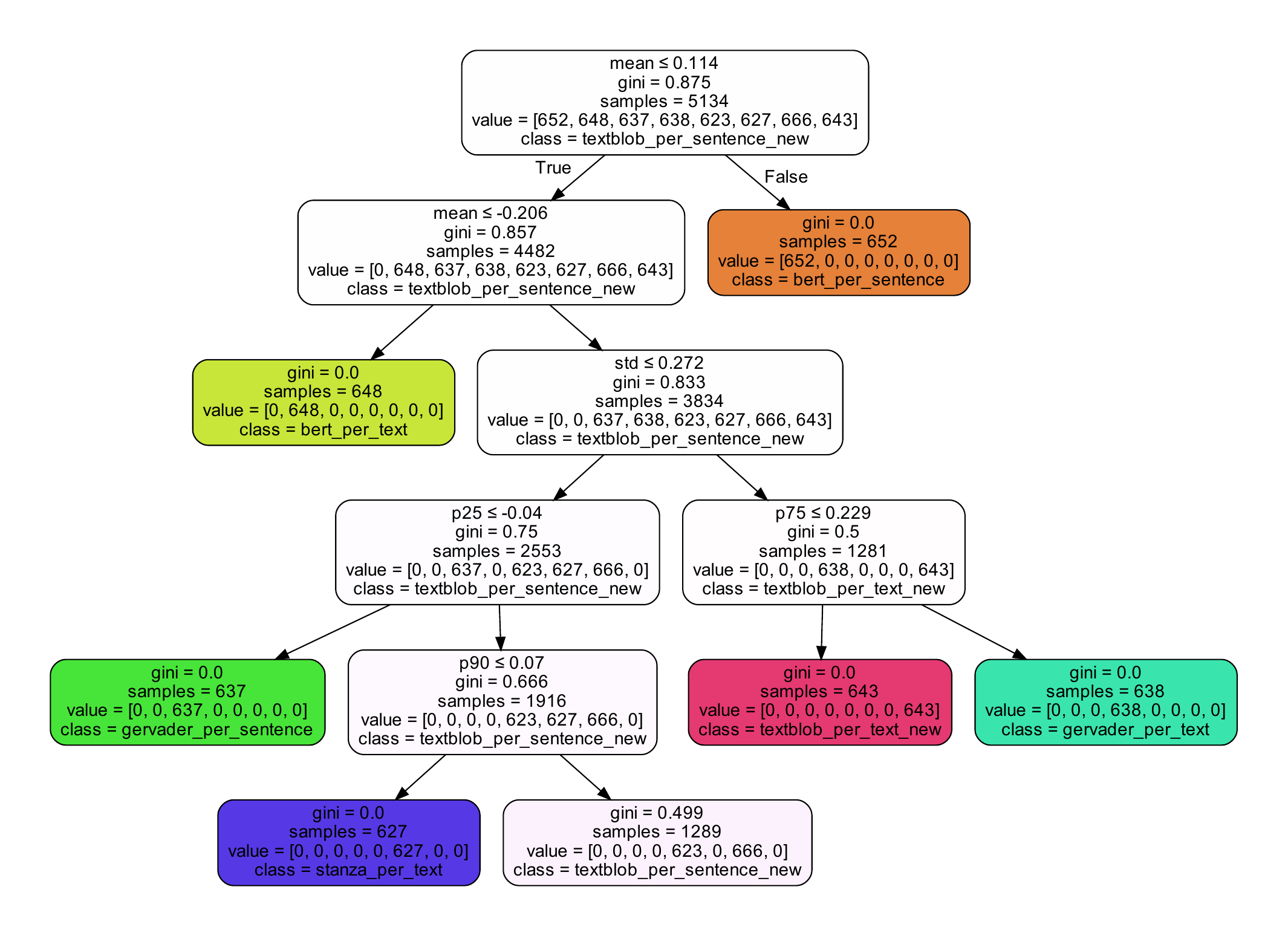}
        \vspace{-0.5cm}
        \caption{DE Decision Tree}
    \end{subfigure}
    \caption{Cutouts of the Decision Trees on the C3 corpus with chunk size $400$. Especially for French and German, very clear rules can be identified. Structure of the node from top to bottom: (1) feature by which the data is divided (2) gini: strength of the feature (3) number of data in the node (4) division into classes (5) class: strongest class in the node (class decision that would be made). More information see \autoref{sec:Other_classifier}.}
    \label{fig:decision_trees_all}
\end{figure*}

\begin{figure*}
    \centering
    \begin{subfigure}[b]{0.90\textwidth}
        \centering
        \includegraphics[width=.95\linewidth, trim={54cm 0 100cm 0},clip]{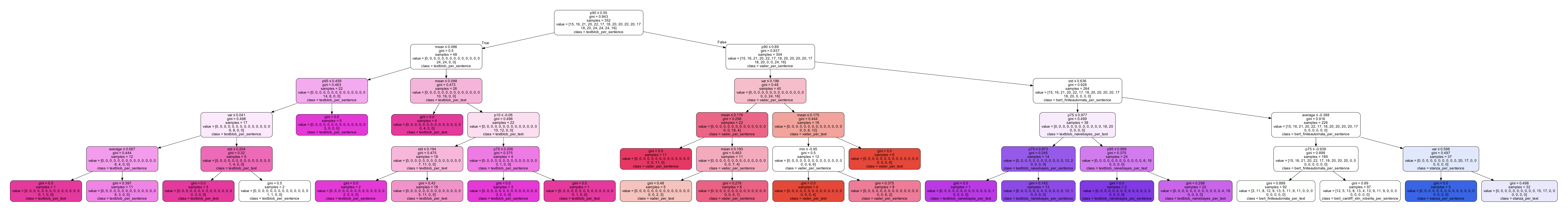}
        \vspace{-0.3cm}
        \caption{EN Decision Tree}
    \end{subfigure}
    \vskip\baselineskip
    \begin{subfigure}[b]{0.90\textwidth}  
        \centering 
        \includegraphics[width=.95\linewidth, trim={13cm 0 9cm 0},clip]{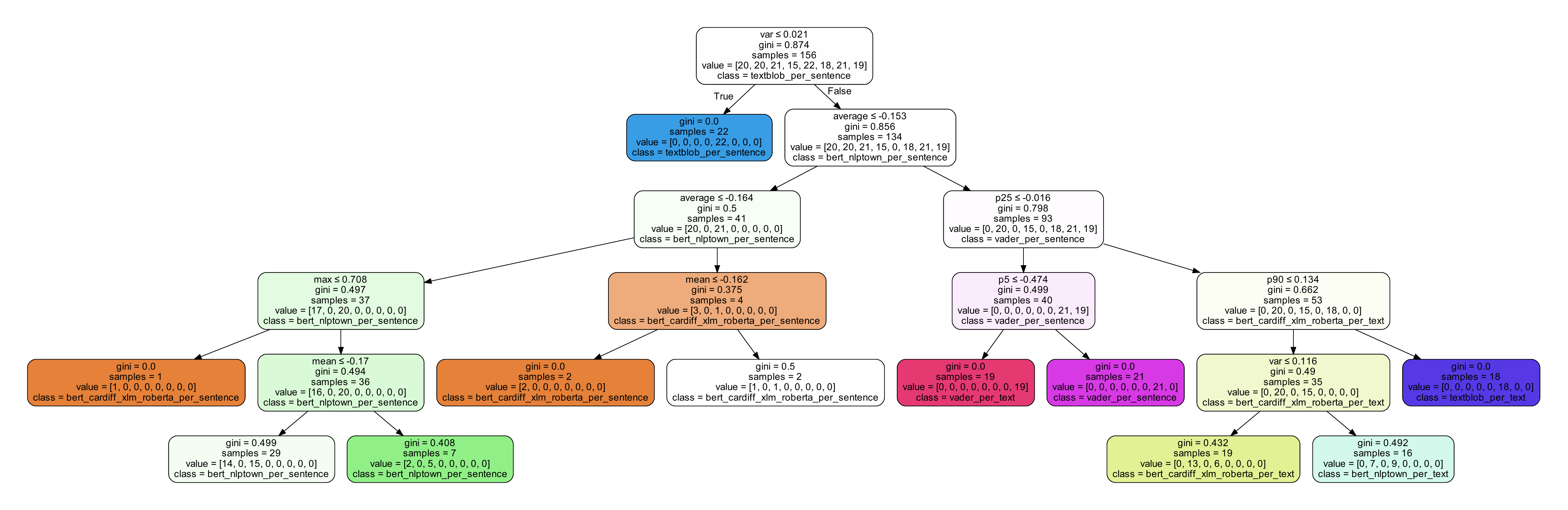}
        \vspace{-0.5cm}
        \caption{FR Decision Tree}
    \end{subfigure}
    \vskip\baselineskip
    \begin{subfigure}[b]{0.85\textwidth}   
        \centering 
        \includegraphics[width=.90\linewidth, trim={0 0 0 0},clip]{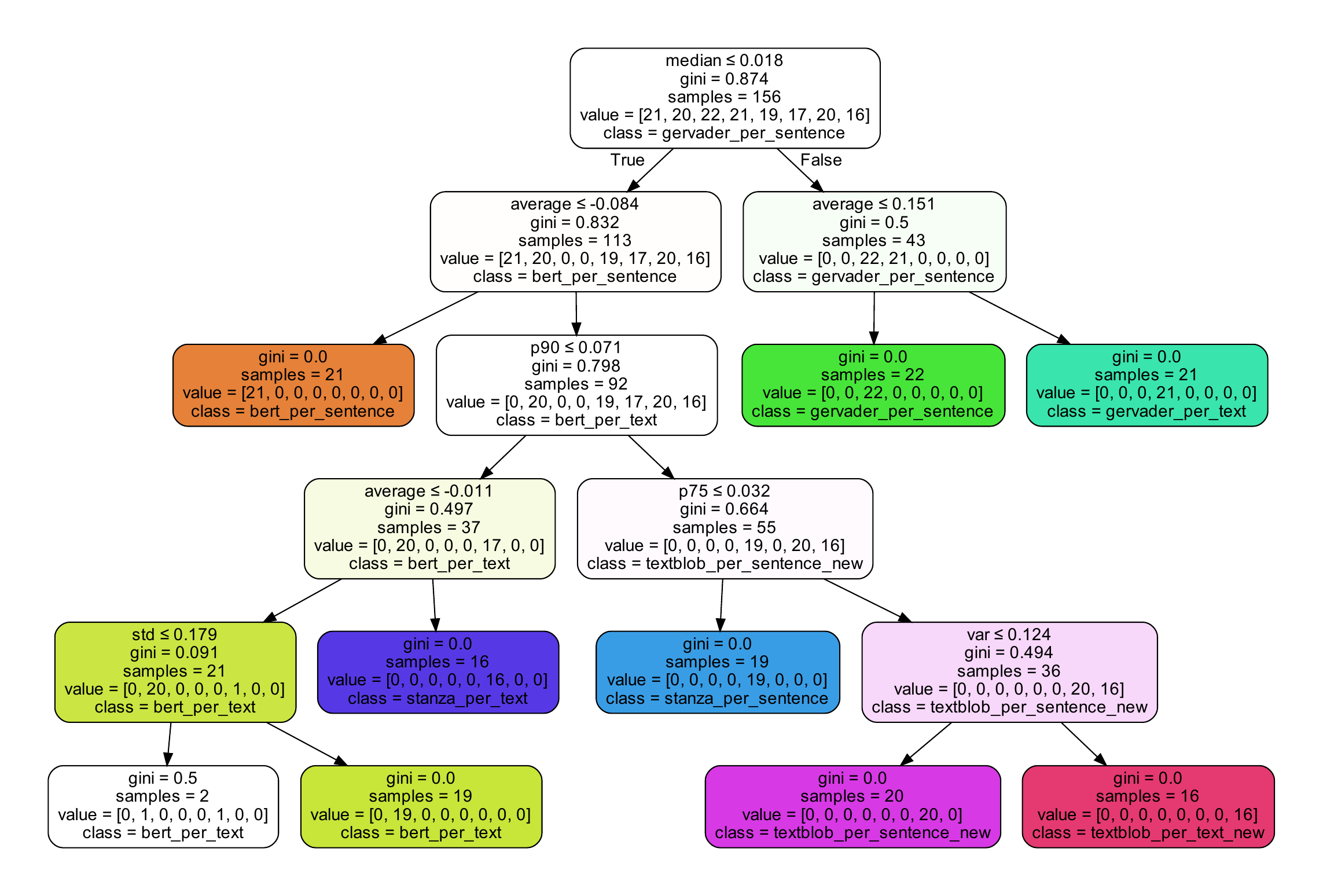}
        \vspace{-0.5cm}
        \caption{DE Decision Tree}
    \end{subfigure}
    \caption{Cutouts of the Decision Trees on the Europarl corpus with chunk size $400$. Especially for German, very clear rules can be identified. Structure of the node from top to bottom: (1) feature by which the data is divided (2) gini: strength of the feature (3) number of data in the node (4) division into classes (5) class: strongest class in the node (class decision that would be made). More information see \autoref{sec:Other_classifier}.}
    \label{fig:decision_trees_eu}
\end{figure*}

\begin{figure*}
    \centering
    \includegraphics[width=1\linewidth]{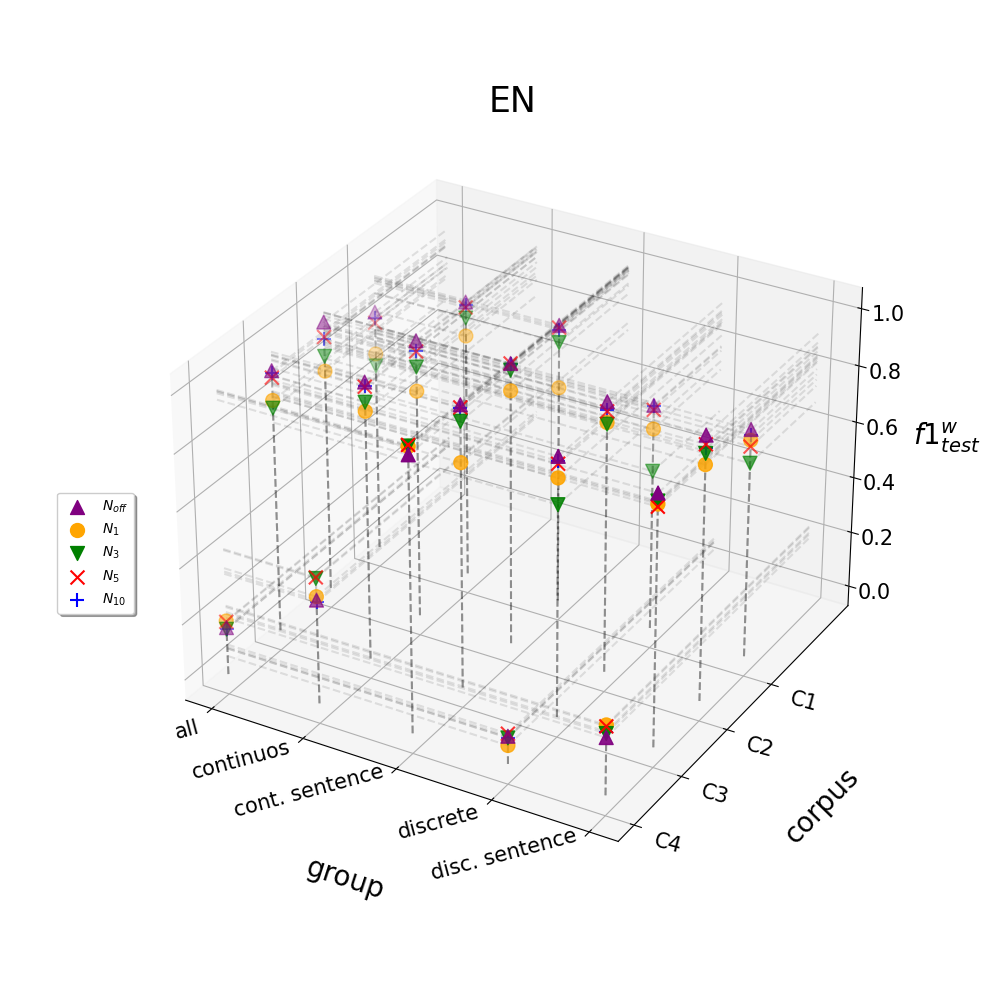}
	\includegraphics[width=0.49\linewidth]{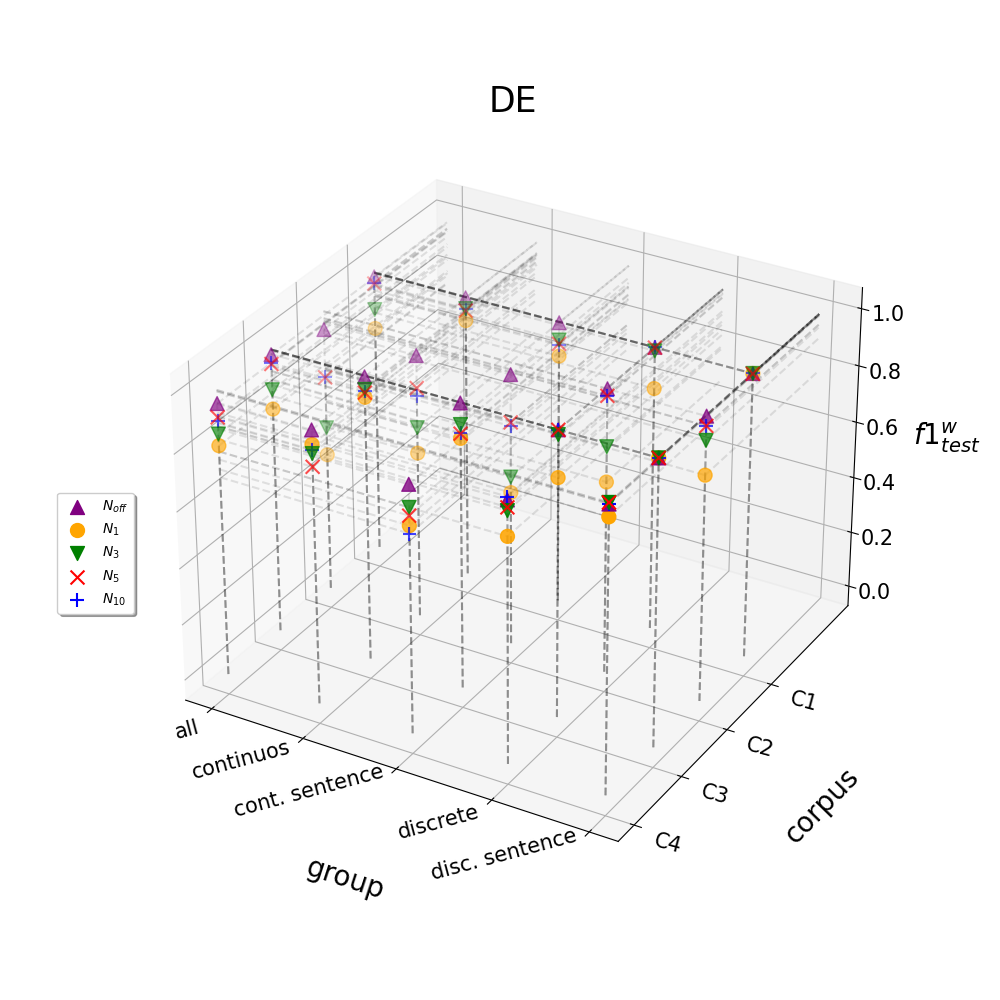}
	\includegraphics[width=0.49\linewidth]{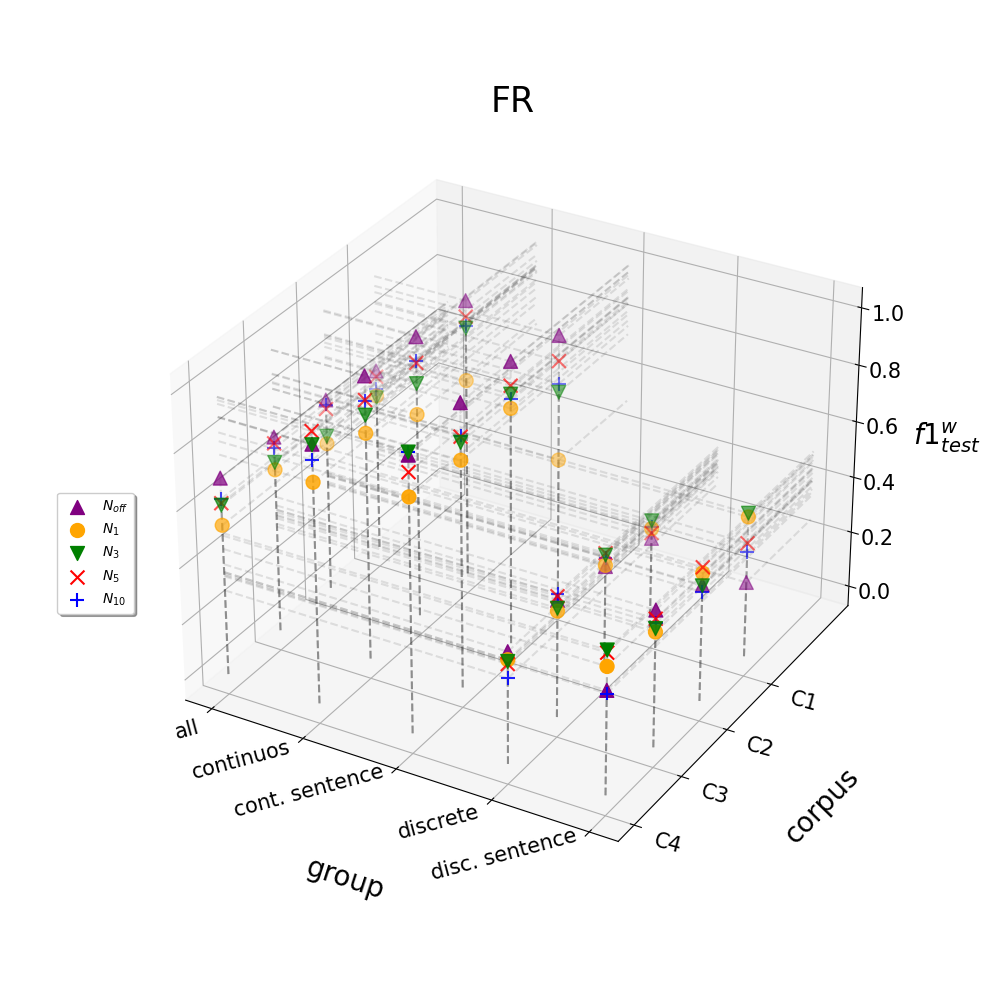}
    \caption{3D visualization of \autoref{tab:big_table_means}, including different normalization results.}
    \label{fig:3d_stats}
\end{figure*}

We provide additional visualizations and data in the appendix on the next pages, mainly showing results on German and French corpora as well as an more detailed overview of the neural network based classifier results on English corpora and the decision trees.
\autoref{tab:corpora_stats} gives an overview of the corpora, with indication of its size and covered time spans.
We show on what data domain the tools we experimented with are trained in \autoref{tab:tools_used_datasets}, which also provides links to the implementation on GitHub and Hugging Face.
Additional visualizations of the German and French corpora are given in the figures~\ref{fig:boxplot_defrwiki_sputnikv}, \ref{fig:dcor_all_de_fr_en}, \ref{fig:voting_de_fr} and \ref{fig:svm_res_all_de_fr_and_ep}.
\autoref{fig:decision_trees_all} and~\ref{fig:decision_trees_eu} show cutouts of the decision trees for all languages.
We add tables showing samples of the experiment results on the English corpora.
That is \autoref{tab:big_table_raw}, which shows macro and weighted $F_1$-scores for the development and test set for a selection of groups and corpora, and \autoref{tab:big_table_means} which provides statistical data about the scores collected over all chunk sizes.
Lastly we provide a 3D visualization of \autoref{tab:big_table_means} in \autoref{fig:3d_stats}.
\begin{table*}
    \centering
    \footnotesize
    \begin{tabular}{lllrrr}
        \toprule
        \textbf{Lang} & \textbf{Corpus} & \textbf{Included in corpus} & \textbf{\# Docs} & \textbf{\# Token} & \textbf{Date} \\
        \midrule
        EN & Wikipedia & C2, C3 & \numprint{21997} & \numprint{10430671} & Dump 2021-06-20 \\
        EN & New York Times & C2, C3 & \numprint{21960} & \numprint{13873994} & 1987 -- 2019 \\
        EN & Europarl & C4 & \numprint{11000} & \numprint{313112} & Release v7 2012 \\
        EN & \#Aufschrei & C1, C3 & \numprint{777} & \numprint{20465} & 2013-06-03 -- 2021-05-21 \\
        EN & \#allemalneschichtmachen & C1, C3 & \numprint{752} & \numprint{7324} & 2021-04-24 -- 2021-04-26 \\
        EN & \#allenichtganzdicht & C1, C3 & \numprint{857} & \numprint{11423} & 2020-04-14 -- 2021-04-23 \\
        EN & \#allesdichtmachen & C1, C3 & \numprint{1135} & \numprint{14807} & 2021-03-11 -- 2021-04-23 \\
        EN & \#lockdownfuerimmer & C1, C3 & \numprint{1498} & \numprint{19726} & 2020-06-23 -- 2021-04-23 \\
        EN & \#niewiederaufmachen & C1, C3 & \numprint{1514} & \numprint{18702} & 2021-04-22 -- 2021-04-23 \\
        EN & \#SputnikV & C1, C3 & \numprint{164504} & \numprint{3332801} & 2020-08-11 -- 2021-05-03 \\
        EN & \#TatortBoykott & C1, C3 & \numprint{164} & \numprint{1123} & 2017-12-15 -- 2021-05-03 \\
        \midrule
        DE & Wikipedia & C2, C3 & \numprint{21999} & \numprint{10707115} & Dump 2021-06-20 \\
        DE & Süddeutsche Zeitung & C2, C3 & \numprint{22001} & \numprint{9365670} & 1992 -- 2014 \\
        DE & Europarl & C4 & \numprint{11000} & \numprint{350864} & Release v7 2012 \\
        DE & \#Aufschrei & C1, C3 & \numprint{197713} & \numprint{5356038} & 2013-06-03 -- 2021-05-21 \\
        DE & \#allemalneschichtmachen & C1, C3 & \numprint{9602} & \numprint{221448} & 2021-04-24 -- 2021-04-26 \\
        DE & \#allenichtganzdicht & C1, C3 & \numprint{36189} & \numprint{836361} & 2020-04-14 -- 2021-04-23 \\
        DE & \#allesdichtmachen & C1, C3 & \numprint{24826} & \numprint{565868} & 2021-03-11 -- 2021-04-23 \\
        DE & \#lockdownfuerimmer & C1, C3 & \numprint{17113} & \numprint{336476} & 2020-06-23 -- 2021-04-23 \\
        DE & \#niewiederaufmachen & C1, C3 & \numprint{19280} & \numprint{394170} & 2021-04-22 -- 2021-04-23 \\
        DE & \#SputnikV & C1, C3 & \numprint{14873} & \numprint{374925} & 2020-08-11 -- 2021-05-03 \\
        DE & \#TatortBoykott & C1, C3 & \numprint{4061} & \numprint{93634} & 2017-12-15 -- 2021-05-03 \\
        \midrule
        FR & Wikipedia & C2, C3 & \numprint{21993} & \numprint{9342397} & Dump 2021-06-20  \\
        FR & Newscrawl 2020 & C2, C3 & \numprint{10000} & \numprint{240548} & 2020\\
        FR & Europarl & C4 & \numprint{11000} & \numprint{354870} & Release v7 2012 \\
        FR & \#SputnikV & C1, C3 & \numprint{14566} & \numprint{411391} & 2020-08-11 -- 2021-05-03 \\
        \bottomrule
    \end{tabular}
    \caption{Overview of the corpora used in our study, the number of documents (articles or tweets), and the date or version in which these data were obtained.}
    \label{tab:corpora_stats}
\end{table*}
\begin{table*}
    \centering
    \resizebox{\linewidth}{!}
    {
        \begin{tabular}{|l|l|c|c|c|c|c|c|c|c}
            \toprule
            \textbf{Tool} & \textbf{Implementation} & \kreis{1} & \kreis{2} & \kreis{3} & \kreis{4} & \kreis{5} & \kreis{6} \\
            \midrule
            TextBlob EN & \quadneu{1} \href{https://github.com/sloria/textblob}{\nolinkurl{sloria/textblob}} & & & & & & \cmark \\
            \midrule
            TextBlob NB EN & \quadneu{1} \href{https://github.com/sloria/textblob}{\nolinkurl{sloria/textblob}} & & & & & & \cmark \\
            \midrule
            TextBlob DE & \quadneu{1} \href{https://github.com/markuskiller/textblob-de}{\nolinkurl{markuskiller/textblob-de}} & & & & & \cmark &  \\
            \midrule
            TextBlog FR & \quadneu{1} \href{https://github.com/sloria/textblob-fr}{\nolinkurl{sloria/textblob-fr}} & & & & & & \cmark \\
            \midrule
            Stanza EN & \quadneu{1} \href{https://github.com/stanfordnlp/stanza}{\nolinkurl{stanfordnlp/stanza}} & \cmark & & \cmark & \cmark&\cmark & \\
            \midrule
            Stanza DE & \quadneu{1} \href{https://github.com/stanfordnlp/stanza}{\nolinkurl{stanfordnlp/stanza}} & \cmark & & & & & \\
            \midrule
            VADER EN & \quadneu{1} \href{https://github.com/cjhutto/vaderSentiment}{\nolinkurl{cjhutto/vaderSentiment}}   &  & &  &  & \cmark & \\
            \midrule
            GerVADER & \quadneu{1} \href{https://github.com/KarstenAMF/GerVADER}{\nolinkurl{KarstenAMF/GerVADER}} & & & & & \cmark & \\
            \midrule
            VADER FR & \quadneu{1} \href{https://github.com/thomas7lieues/vader_FR}{\nolinkurl{thomas7lieues/vader_FR}} &   & &  &  & \cmark  & \\
            \midrule
            twitter-roberta & \quadneu{2} \href{https://huggingface.co/cardiffnlp/twitter-roberta-base-sentiment}{\nolinkurl{cardiffnlp/twitter-roberta-base-sentiment}} & \cmark & & & & & \\
            \midrule
            twitter-xlm-roberta & \quadneu{2} \href{https://huggingface.co/cardiffnlp/twitter-xlm-roberta-base-sentiment}{\nolinkurl{cardiffnlp/twitter-xlm-roberta-base-sentiment}} & \cmark & & & & & \\
            \midrule
            finiteautomata & \quadneu{2} \href{https://huggingface.co/finiteautomata/bertweet-base-sentiment-analysis}{\nolinkurl{finiteautomata/bertweet-base-sentiment-analysis}} & \cmark & & & & & \\
            \midrule
            nlptown & \quadneu{2} \href{https://huggingface.co/nlptown/bert-base-multilingual-uncased-sentiment}{\nolinkurl{nlptown/bert-base-multilingual-uncased-sentiment}} & & & \cmark & & & \\
            \midrule
            german-sentiment & \quadneu{2} \href{https://huggingface.co/oliverguhr/german-sentiment-bert}{\nolinkurl{oliverguhr/german-sentiment-bert}}  & \cmark & \cmark & \cmark & \cmark & & \\
            \midrule
            siebert & \quadneu{2} \href{https://huggingface.co/siebert/sentiment-roberta-large-english}{\nolinkurl{siebert/sentiment-roberta-large-english}} & \cmark &   &\cmark & & & \\
            \bottomrule
        \end{tabular}
    }
    \caption{Overview of the datasets grouped by domain used by the selected sentiment detection tools and the implementation used by our experiments.
    The data is grouped in the following categories:
    \kreis{1} Social media posts like Twitter, \kreis{2} specialised texts based on books or Wikipedia, \kreis{3} text taken from ratings or reviews, \kreis{4} emoticons, \kreis{5} entries from semantic lexicons and \kreis{6} unknown training data.
    \quadneu{1} denotes a link to GitHub and \quadneu{2} to the Hugging Face repositories.}
    \label{tab:tools_used_datasets}
\end{table*}
\begin{table*}
    \centering
    \small
    \begin{tabularx}{\linewidth}{XXrrrrrrr}
        \toprule
        \textbf{Group} & \textbf{Corpus} & \textbf{Chunks} & \textbf{Dev m $F_1$} & \textbf{Dev w $F_1$} & \textbf{Test m $F_1$} & \textbf{Test w $F_1$} & \textbf{\# Train} & \textbf{\# Test} \\  
        \midrule
all & C3 & 50 & 0.869 & 0.868 & 0.868 & 0.867 & \numprint{54116} & \numprint{11597} \\
all & C3 & 70 & 0.888 & 0.889 & 0.891 & 0.889 & \numprint{38656} & \numprint{8284} \\
all & C3 & 90 & 0.899 & 0.898 & 0.893 & 0.894 & \numprint{30076} & \numprint{6445} \\
all & C3 & 100 & 0.909 & 0.909 & 0.909 & 0.908 & \numprint{27064} & \numprint{5800} \\
all & C3 & 200 & 0.937 & 0.941 & 0.94 & 0.939 & \numprint{13532} & \numprint{2900} \\
all & C3 & 400 & 0.961 & 0.961 & 0.962 & 0.96 & \numprint{6766} & \numprint{1450} \\
all & C3 & 800 & 0.991 & 0.992 & 0.974 & 0.972 & \numprint{3388} & \numprint{727} \\
all & C3 & 1000 & 0.996 & 0.997 & 0.987 & 0.988 & \numprint{2708} & \numprint{581} \\
        \midrule
all & C1 & 50 & 0.76 & 0.759 & 0.753 & 0.751 & \numprint{43054} & \numprint{9226} \\ 
all & C1 & 100 & 0.814 & 0.814 & 0.817 & 0.825 & \numprint{21532} & \numprint{4615} \\
all & C1 & 1000 & 0.975 & 0.974 & 0.958 & 0.959 & \numprint{2154} & \numprint{462} \\
        \midrule
all & C2 & 50 & 0.921 & 0.921 & 0.909 & 0.907 & \numprint{11074} & \numprint{2374} \\
all & C2 & 100 & 0.948 & 0.947 & 0.955 & 0.957 & \numprint{5544} & \numprint{1188} \\
all & C2 & 1000 & 1.0 & 1.0 & 0.978 & 0.974 & \numprint{554} & \numprint{119} \\
        \midrule
all & C4 & 50 & 0.238 & 0.25 & 0.195 & 0.198 & \numprint{2772} & \numprint{594} \\
all & C4 & 100 & 0.223 & 0.235 & 0.163 & 0.174 & \numprint{1386} & \numprint{297} \\
all & C4 & 1000 & 0.243 & 0.317 & 0.092 & 0.074 & \numprint{138} & \numprint{30} \\ 
        \midrule
discrete & C1 & 50 & 0.668 & 0.667 & 0.665 & 0.666 & \numprint{28702} & \numprint{6151} \\ 
discrete & C1 & 1000 & 0.968 & 0.968 & 0.968 & 0.968 & \numprint{1436} & \numprint{308} \\
discrete & C2 & 50 & 0.909 & 0.907 & 0.906 & 0.91 & \numprint{7382} & \numprint{1583} \\ 
discrete & C2 & 1000 & 1.0 & 1.0 & 0.972 & 0.975 & \numprint{368} & \numprint{80} \\
discrete & C3 & 50 & 0.839 & 0.838 & 0.841 & 0.84 & \numprint{36077} & \numprint{7731} \\ 
discrete & C3 & 1000 & 0.981 & 0.979 & 0.993 & 0.992 & \numprint{1806} & \numprint{387} \\
discrete & C4 & 50 & 0.127 & 0.122 & 0.128 & 0.126 & \numprint{1848} & \numprint{396} \\ 
discrete & C4 & 1000 & 0.17 & 0.136 & 0.078 & 0.086 & \numprint{92} & \numprint{20} \\
        \midrule
continuous & C1 & 50 & 0.953 & 0.953 & 0.952 & 0.952  & \numprint{14350} & \numprint{3076} \\ 
continuous & C1 & 1000 & 1.0 & 1.0 & 1.0 & 1.0 & \numprint{718} & \numprint{154} \\
continuous & C2 & 50 & 0.973 & 0.971 & 0.952 & 0.954  & \numprint{3691} & \numprint{792} \\ 
continuous & C2 & 1000 & 1.0 & 1.0 & 0.982 & 0.975 & \numprint{184} & \numprint{40} \\
continuous & C3 & 50 & 0.943 & 0.942 & 0.943 & 0.945 & \numprint{18038} & \numprint{3866} \\ 
continuous & C3 & 1000 & 1.0 & 1.0 & 0.994 & 0.995 & \numprint{902} & \numprint{194} \\
continuous & C4 & 50 & 0.506 & 0.508 & 0.472 & 0.463  & \numprint{924} & \numprint{198} \\ 
continuous & C4 & 1000 & 0.511 & 0.627 & 0.25 & 0.15 & \numprint{46} & \numprint{10} \\
        \midrule
disc.~sentences & C1 & 50 & 0.666 & 0.667 & 0.677 & 0.675  & \numprint{14350} & \numprint{3076} \\ 
disc.~sentences & C1 & 1000 & 0.979 & 0.981 & 0.955 & 0.954 & \numprint{718} & \numprint{154} \\ 
disc.~sentences & C2 & 50 & 0.884 & 0.886 & 0.867 & 0.873  & \numprint{3691} & \numprint{791} \\ 
disc.~sentences & C2 & 1000 & 1.0 & 1.0 & 0.943 & 0.923 & \numprint{230} & \numprint{50} \\
disc.~sentences & C3 & 50 & 0.804 & 0.806 & 0.8 & 0.8  & \numprint{18038} & \numprint{3866} \\ 
disc.~sentences & C3 & 1000 & 0.974 & 0.975 & 0.975 & 0.974 & \numprint{902} & \numprint{194} \\
disc.~sentences & C4 & 50 & 0.291 & 0.277 & 0.267 & 0.266  & \numprint{924} & \numprint{198} \\ 
disc.~sentences & C4 & 1000 & 0.337 & 0.337 & 0.167 & 0.2 & \numprint{46} & \numprint{10} \\
        \midrule
cont.~sentences & C1 & 50 & 0.961 & 0.962 & 0.968 & 0.968 & \numprint{7175} & \numprint{1538} \\ 
cont.~sentences & C1 & 1000 & 1.0 & 1.0 & 1.0 & 1.0 & \numprint{359} & \numprint{77} \\
cont.~sentences & C2 & 50 & 1.0 & 1.0 & 0.995 & 0.995 & \numprint{1845} & \numprint{396} \\ 
cont.~sentences & C2 & 1000 & 1.0 & 1.0 & 0.943 & 0.949 & \numprint{92} & \numprint{20} \\
cont.~sentences & C3 & 50 & 0.972 & 0.972 & 0.968 & 0.969 & \numprint{9019} & \numprint{1933} \\ 
cont.~sentences & C3 & 1000 & 1.0 & 1.0 & 1.0 & 1.0 & \numprint{451} & \numprint{97} \\
cont.~sentences & C4 & 50 & 1.0 & 1.0 & 1.0 & 1.0 & \numprint{462} & \numprint{99} \\ 
cont.~sentences & C4 & 1000 & 1.0 & 1.0 & 1.0 & 1.0 & \numprint{23} & \numprint{5} \\
        \bottomrule
    \end{tabularx}
    \caption{This table shows a selection of the raw results produced by the NN experiment described in \autoref{sec:nn}.
    The selection contains non-normalized data for different corpora, groupings and chunk sizes.
    We provide macro and weighted $F_1$-scores for the test and development set calculated by scikit-learn.}
    \label{tab:big_table_raw}
\end{table*}
%


\begin{table*}
    \centering
    \footnotesize
    \begin{tabularx}{\linewidth}{Xlrrrrrr}
        \toprule
        \textbf{Group} & \textbf{Corpus} & \textbf{Dev mean $F_1$} & \textbf{Dev std $F_1$} & \textbf{Test mean $F_1$} & \textbf{Test std $F_1$} & \textbf{Test min $F_1$} & \textbf{Test max $F_1$} \\ 
        \midrule 
all & C3 & $0.932$ & $0.048$ & $0.927$ & $0.044$ & $0.867$ & $0.988$ \\ 
all & C1 & $0.865$ & $0.087$ & $0.86$ & $0.079$ & $0.751$ & $0.959$ \\ 
all & C2 & $0.964$ & $0.033$ & $0.959$ & $0.033$ & $0.907$ & $1.0$ \\ 
all & C4 & $0.253$ & $0.031$ & $0.169$ & $0.048$ & $0.074$ & $0.227$ \\ 
        \midrule
discrete & C3 & $0.912$ & $0.058$ & $0.911$ & $0.055$ & $0.84$ & $0.992$ \\ 
discrete & C1 & $0.812$ & $0.118$ & $0.8$ & $0.114$ & $0.666$ & $0.968$ \\ 
discrete & C2 & $0.964$ & $0.035$ & $0.953$ & $0.028$ & $0.91$ & $0.99$ \\ 
discrete & C4 & $0.136$ & $0.013$ & $0.099$ & $0.032$ & $0.029$ & $0.132$ \\ 
        \midrule
continuous & C3 & $0.978$ & $0.023$ & $0.978$ & $0.02$ & $0.945$ & $1.0$ \\ 
continuous & C1 & $0.981$ & $0.018$ & $0.982$ & $0.018$ & $0.952$ & $1.0$ \\ 
continuous & C2 & $0.992$ & $0.011$ & $0.982$ & $0.017$ & $0.954$ & $1.0$ \\ 
continuous & C4 & $0.555$ & $0.045$ & $0.37$ & $0.106$ & $0.15$ & $0.478$ \\ 
        \midrule
disc.~sentences & C3 & $0.886$ & $0.063$ & $0.883$ & $0.063$ & $0.8$ & $0.974$ \\ 
disc.~sentences & C1 & $0.807$ & $0.112$ & $0.808$ & $0.118$ & $0.675$ & $0.969$ \\ 
disc.~sentences & C2 & $0.959$ & $0.044$ & $0.932$ & $0.044$ & $0.873$ & $1.0$ \\ 
disc.~sentences & C4 & $0.29$ & $0.036$ & $0.204$ & $0.099$ & $0.038$ & $0.384$ \\ 
        \midrule
cont.~sentences & C3 & $0.992$ & $0.01$ & $0.992$ & $0.011$ & $0.969$ & $1.0$ \\ 
cont.~sentences & C1 & $0.99$ & $0.013$ & $0.988$ & $0.014$ & $0.968$ & $1.0$ \\ 
cont.~sentences & C2 & $1.0$ & $0.0$ & $0.993$ & $0.018$ & $0.949$ & $1.0$ \\ 
cont.~sentences & C4 & $1.0$ & $0.0$ & $0.967$ & $0.053$ & $0.857$ & $1.0$ \\ 
        \bottomrule
    \end{tabularx}
    \caption{Sample of the results of the experiments on the EN corpora from experiment in \autoref{sec:nn}.
    The results show the non-normalized experiments for all corpora and groupings, the weighted $F_1$-scores, as reported by scikit-mean, are averaged over the different chunk sizes.}
    \label{tab:big_table_means}
\end{table*}
%

%

\end{document}